%% file: main.tex
\pdfoutput=1
\documentclass[10pt,twocolumn,letterpaper]{article}

\usepackage[pagenumbers]{cvpr} 

\input{preamble}

\definecolor{cvprblue}{rgb}{0.21,0.49,0.74}
\usepackage[pagebackref,breaklinks,colorlinks,allcolors=cvprblue]{hyperref}

\def\paperID{14798} 
\def\confName{CVPR}
\def\confYear{2025}

\newcommand{\firstauth}{\textsuperscript{†}}
\newcommand{\corrauth}{\textsuperscript{*}}

\title{Adaptive Rectangular Convolution for Remote Sensing Pansharpening}

\author{
Xueyang Wang\firstauth, 
Zhixin Zheng, 
Jiandong Shao, 
Yule Duan, 
Liang-Jian Deng\corrauth \\ 
University of Electronic Science and Technology of China, Chengdu, China \\[-2pt] 
{\tt\small xywang\_uestc@std.uestc.edu.cn; zhengzx@std.uestc.edu.cn; shamy@std.uestc.edu.cn} \\[-2pt] 
{\tt\small yule.duan@outlook.com; liangjian.deng@uestc.edu.cn} 
}

\begin{document}
\maketitle

\renewcommand{\thefootnote}{}
\footnotetext{
  \noindent
  \firstauth First author.\\
  \corrauth Corresponding author.
}

\renewcommand{\thefootnote}{\arabic{footnote}} 
\input{sec/0_abstract}    
\input{sec/1_intro}
\input{sec/2_related_work}
\input{sec/3_methods}
\input{sec/4_experiments}

\input{sec/5_conclusion}

\bibliographystyle{plain}
\bibliography{main}
\input{sec/X_suppl}

\end{document}

%% file: preamble.tex
\usepackage[dvipsnames]{xcolor}
\usepackage{booktabs}
\usepackage{multirow}
\usepackage{adjustbox}
\usepackage{tcolorbox}
\usepackage{tabularx}
\usepackage{enumitem} 
\usepackage{subcaption}
\usepackage{dashrule} 
\usepackage{float}
\setlist[enumerate]{itemsep=0.1em, topsep=0.5em} 


%% file: sec/0_abstract.tex
\begin{abstract}
Recent advancements in convolutional neural network (CNN)-based techniques for remote sensing pansharpening have markedly enhanced image quality. However, conventional convolutional modules in these methods have two critical drawbacks. First, the sampling positions in convolution operations are confined to a fixed square window. Second, the number of sampling points is preset and remains unchanged. Given the diverse object sizes in remote sensing images, these rigid parameters lead to suboptimal feature extraction. To overcome these limitations, we introduce an innovative convolutional module, Adaptive Rectangular Convolution (ARConv). ARConv adaptively learns both the height and width of the convolutional kernel and dynamically adjusts the number of sampling points based on the learned scale. This approach enables ARConv to effectively capture scale-specific features of various objects within an image, optimizing kernel sizes and sampling locations. Additionally, we propose ARNet, a network architecture in which ARConv is the primary convolutional module. Extensive evaluations across multiple datasets reveal the superiority of our method in enhancing pansharpening performance over previous techniques. Ablation studies and visualization further confirm the efficacy of ARConv. The source code will be available at  \url{https://github.com/WangXueyang-uestc/ARConv.git}.
\end{abstract}

%% file: sec/1_intro.tex
\section{Introduction}
\label{sec:intro}

\begin{figure}[htbp]
  \centering
  \includegraphics[width=\linewidth, trim=240 20 200 10, clip]{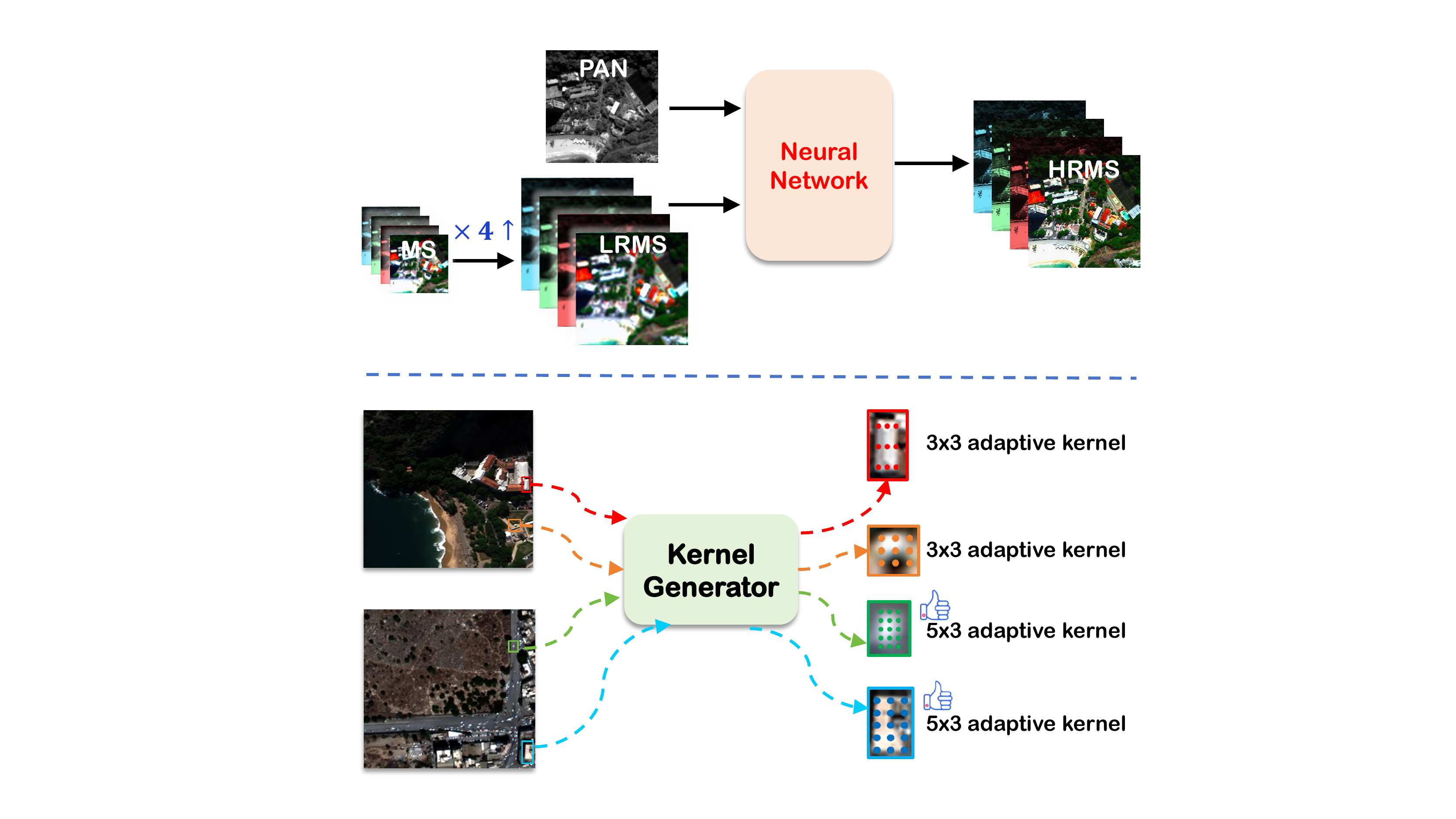} 
  \caption{Top row: The comprehensive flowchart of remote sensing pansharpening via a DL-based approach. Bottom row: An illustrative example of our Adaptive Rectangular Convolution (ARConv), boasting two distinct advantages: 1) its convolution kernels can adaptively modify sampling positions in accordance with object sizes; 2) the quantity of sampling points is dynamically determined across various feature maps, for instance, achieving a $5\times 3$ adaptive rectangular convolution, which, to our knowledge, is the first attempt.}
  \label{fig:toyexample}
  \vspace{-0.5cm}
\end{figure}

Clear remote sensing images are critically important in various domains, including military applications and agriculture. However, existing technologies are only capable of capturing low-resolution multispectral images (LRMS) and high-resolution panchromatic images (PAN). LRMS provides rich spectral information but suffers from low spatial resolution, while PAN images, though rich in spatial detail, are limited to grayscale and lack spectral information. The objective of pansharpening is to integrate these two types of images to produce high-resolution multispectral images (HRMS), as shown in \cref{fig:toyexample}. Many pansharpening methods have been proposed \cite{Meng2019ReviewOT}, including traditional methods and deep learning-based methods, among which traditional methods are further divided into Component Substitution (CS) \cite{CS1,CS2}, Multi-Resolution Analysis (MRA) \cite{MRA1, Vivone2018FullSR}, and variational optimization (VO) \cite{VO1, Tian2021VariationalPB}. Recently, the remarkable advancements in deep learning within image processing have led to the widespread application of numerous methods based on convolutional neural networks for pansharpening. Compared to traditional pansharpening methods, the feature of the input PAN and LRMS images in these methods is mainly extracted through convolutional kernels.  However, standard convolution has two major drawbacks. First, its sampling positions are fixed within a square window of a determined size, which restricts its ability to deform, thereby preventing it from adaptively finding the sampling locations. Second, the number of sampling points of the convolutional kernel is predetermined, making it challenging to adaptively capture features at different scales. In remote sensing images, the scale differences between different objects can be significant, such as small cars and large buildings, which standard convolutions are not adept at capturing, leading to inefficient feature extraction.

In recent years, many innovative convolutional methods have been proposed for pansharpening. Spatial adaptive convolution methods, such as PAC \cite{Su2019PixelAdaptiveCN}, DDF \cite{Zhou2021DecoupledDF}, LAGConv \cite{Jin2022LAGConvLA}, and CANConv \cite{Duan2024ContentAdaptiveNC}, can adaptively generate different convolution kernel parameters based on various spatial locations, enabling them to accommodate different spatial regions. However, these methods have yet to fully consider the rich scale information present in remote sensing images. Shape-adaptive convolutions, such as Deformable Convolution \cite{Dai2017DeformableCN, Zhu2018DeformableCV}, can adaptively adjust the position of each sampling point by learning offsets to extract features of objects with different shapes. Although this provides significant flexibility, the number of learnable parameters increases quadratically with the kernel size, making it difficult to achieve convergence on small datasets, such as in image sharpening tasks. Furthermore, it cannot adjust the number of sampling points based on the shape of the convolution kernel, which further limits its performance. Multi-scale convolutions, such as pyramid convolution \cite{PYconv}, can extract information at different scales within the same feature map. \textit{However, the size of their convolution kernels is predetermined, while the features in the image may exhibit different patterns and structures across scales. This can lead to imprecise feature fusion between scales, potentially affecting the model's overall performance.}

Based on the above analysis, we propose the Adaptive Rectangular Convolution (ARConv), which can not only adaptively adjust the sampling positions but also the number of sampling points, as shown in \cref{fig:toyexample}. The former is achieved by learning only two parameters: the height and width of the convolution kernel, without incurring additional computational burden as the kernel size increases. The latter selects an appropriate number of sampling points based on the average level of the learned height and width. Moreover, we introduce affine transformation to ARConv, which brings spatial adaptability. All of this enables our module to effectively extract features from objects of varying sizes in the feature map. The main contributions of this paper are outlined as follows:
\begin{enumerate}
\item ARConv is proposed as a module that can adaptively adjust the sampling positions and change the number of sampling points, enabling it to effectively capture scale-specific features of various objects in remote sensing images. Based on ARConv and the U-net architecture \cite{Unet1, Wang2021SSconvES}, ARNet is introduced.
\item The relationship between the learned height and width of the convolution kernel and the actual object sizes is explored through heatmap visualizations. A certain level of correlation is observed, which validates the effectiveness of the proposed method.
\item The effectiveness of ARConv is validated by comparing it with various pansharpening methods across multiple datasets. Results demonstrate that ARConv achieves outstanding performance.
\end{enumerate}

%% file: sec/2_related_work.tex
\section{Related Works}
\label{sec:relaw}

\begin{figure}[htbp]
    \centering
     \begin{subfigure}{0.49\linewidth}
        \centering
        \setlength{\abovecaptionskip}{-0.35cm} 
        \includegraphics[width=\linewidth, trim=0 0 0 0, clip]{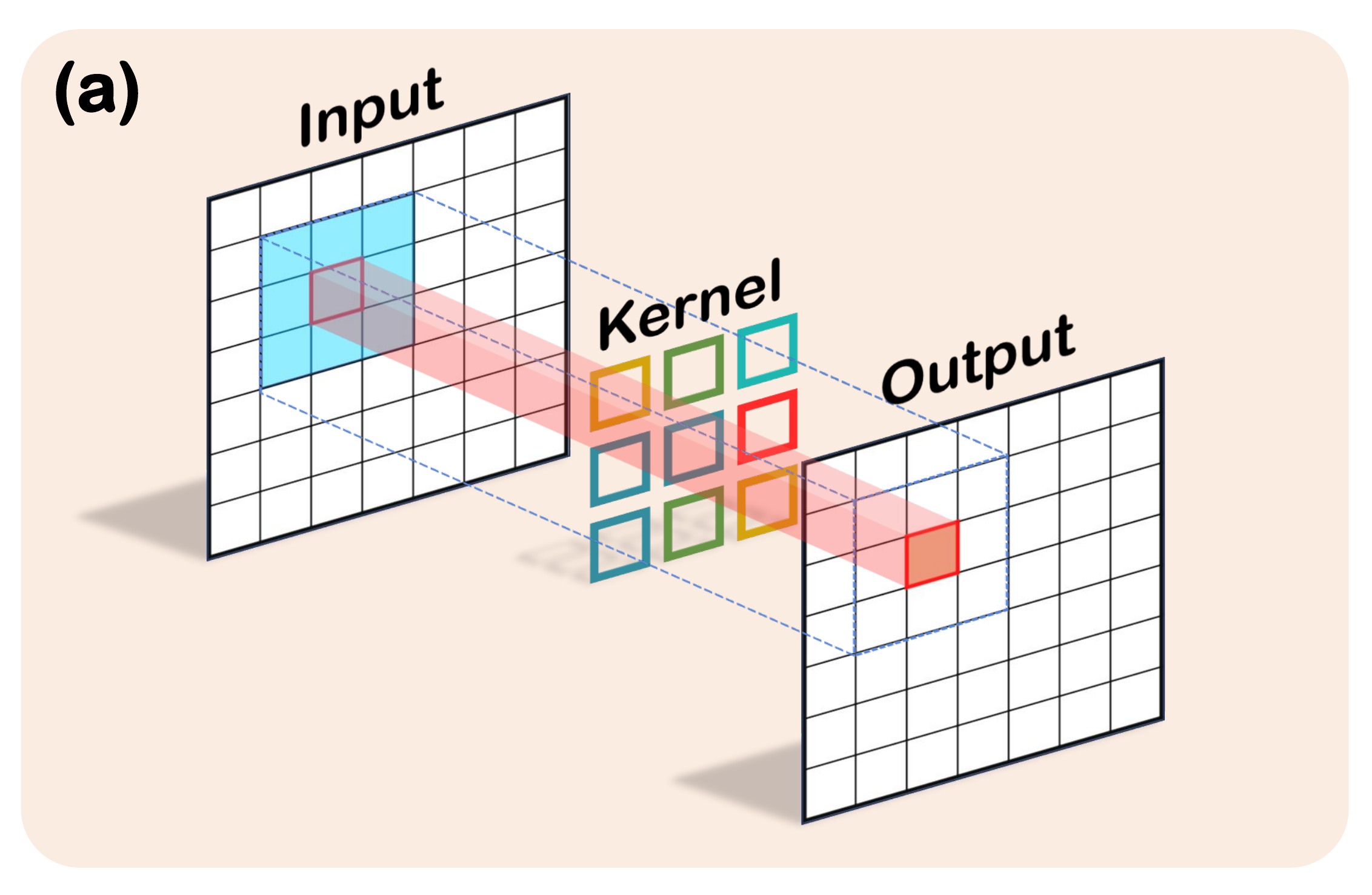}
        \caption*{}
        \label{fig:standordconv}
    \end{subfigure}
    \begin{subfigure}{0.49\linewidth}
        \centering
        \setlength{\abovecaptionskip}{-0.35cm} 
        \includegraphics[width=\linewidth, trim=0 0 0 0, clip]{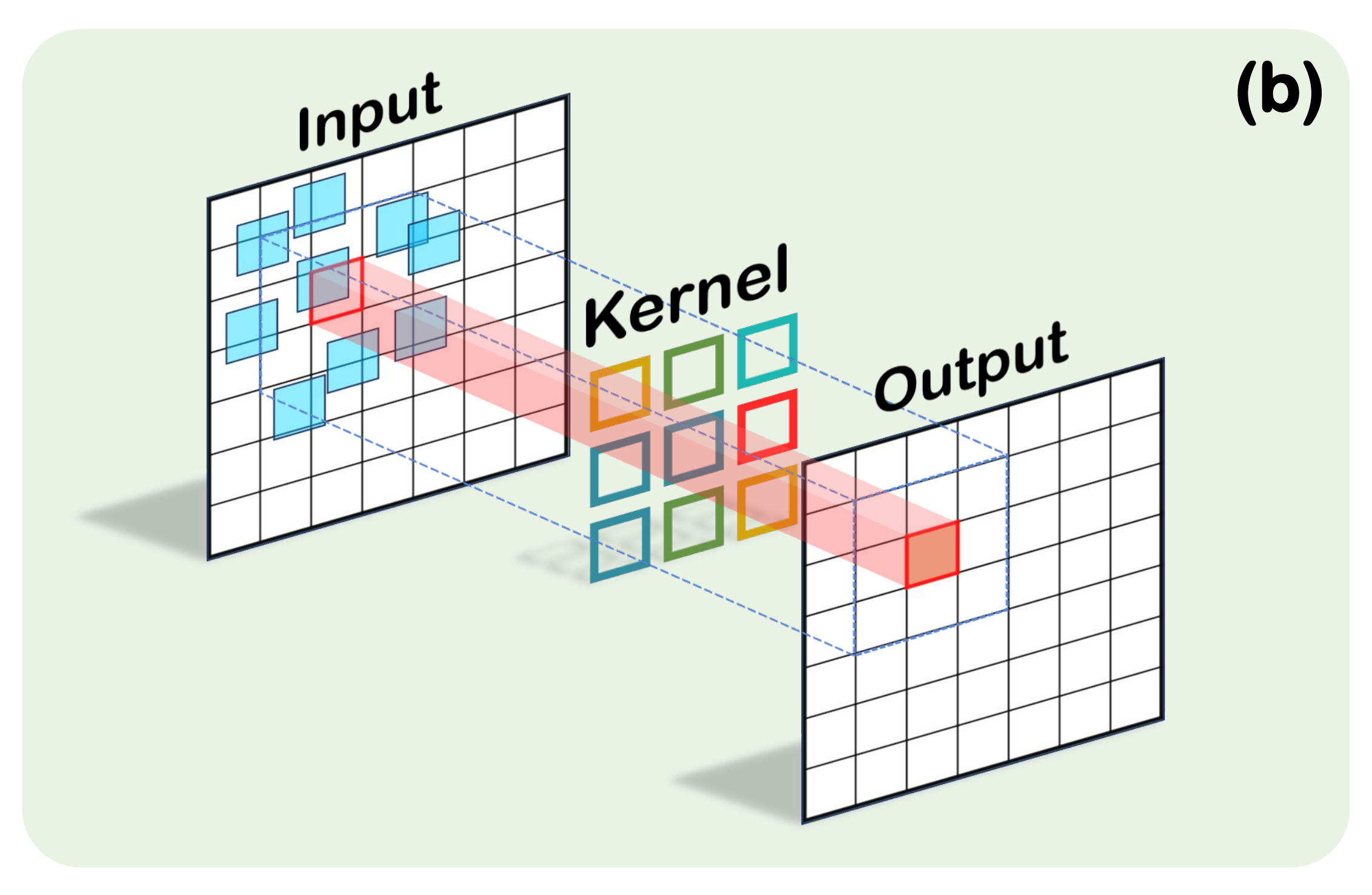}
        \caption*{}
        \label{fig:dcn}
    \end{subfigure}
    \begin{subfigure}{0.49\linewidth}
        \centering
        \setlength{\abovecaptionskip}{-0.35cm} 
        \includegraphics[width=\linewidth, trim=0 0 0 0, clip]{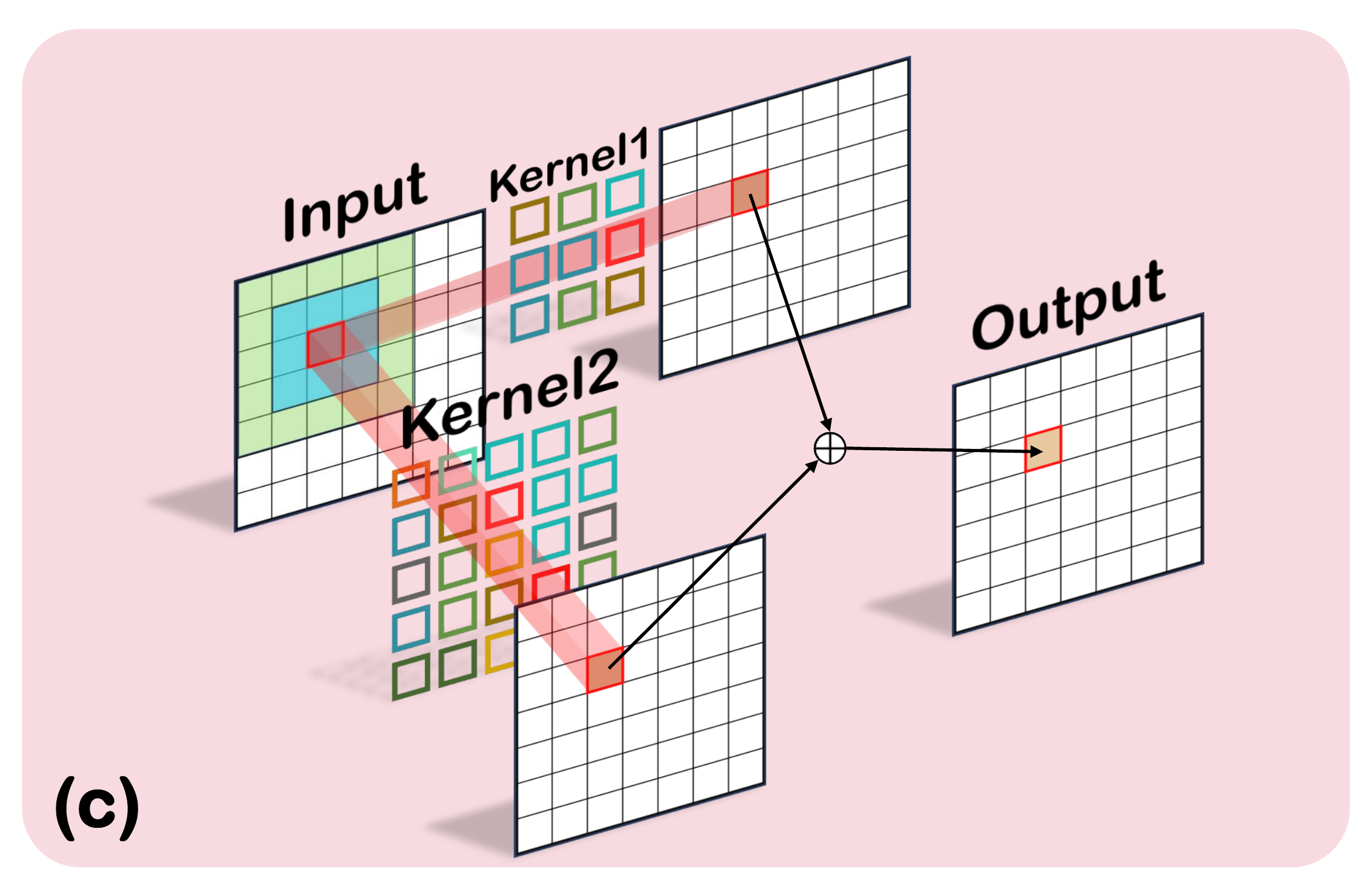}
        \caption*{}
        \label{fig:pyconv}
    \end{subfigure}
    \begin{subfigure}{0.49\linewidth}
        \centering
        \setlength{\abovecaptionskip}{-0.35cm} 
        \includegraphics[width=\linewidth, trim=0 0 0 0, clip]{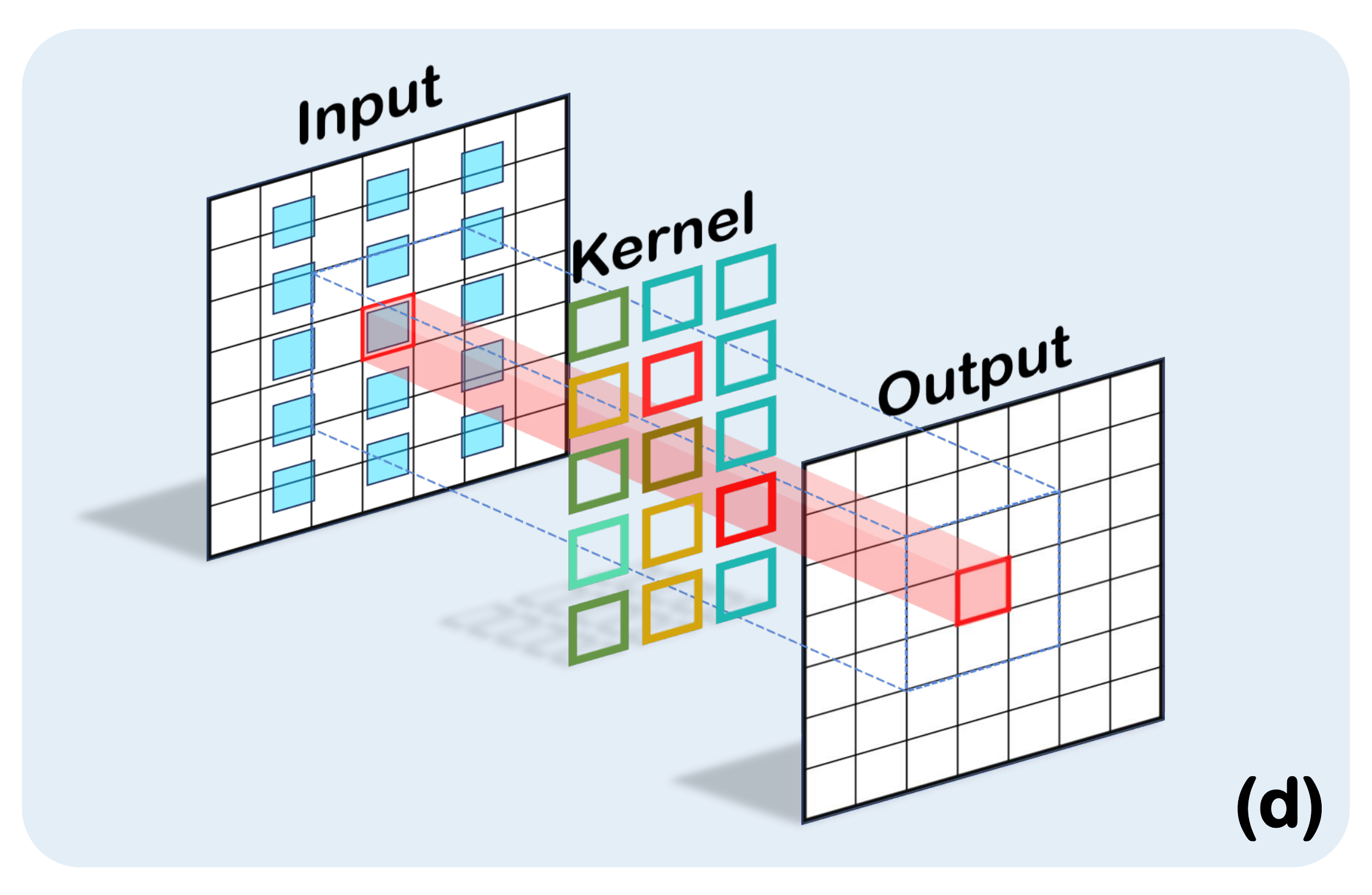}
        \caption*{}
        \label{fig:proposed}
    \end{subfigure}

    \caption{Diagrams illustrating the working principles of four types of convolutional kernels. (a) Standard Convolution. (b) Deformable Convolution \cite{Dai2017DeformableCN,Zhu2018DeformableCV}. (c) Multi-scale Convolution \cite{PYconv,Li2019SelectiveKN}. (d) Our proposed Convolution (ARConv).}
    \vspace{-0.5cm}
    \label{fig:convs}
\end{figure}

\subsection{Adaptive Convolution}
Standard convolution, with its fixed shape and size, exhibits limited flexibility in handling geometric transformations, posing challenges in adapting to the varying scales and shapes of objects commonly encountered in visual tasks. Deformable Convolution \cite{Dai2017DeformableCN, Zhu2018DeformableCV} was the pioneering approach to address this limitation by learning an offset matrix that adjusts the sampling positions of each pixel, as visualized in \cref{fig:convs}. This advancement enabled the convolutional kernel to deform in an unsupervised manner for the first time. Building on the concept of deformable convolution, Dynamic Snake Convolution \cite{snackConvolution} specifically optimizes feature extraction for tubular structures by employing carefully designed loss constraints to guide the deformation of convolutional kernels. Scale-adaptive Convolution \cite{scale-adaptive} extends this flexibility by allowing the convolutional kernel to learn scaling ratios, dynamically modifying the receptive field to better capture features at different scales.

In the convolutions described above, the deformation either becomes overly flexible, leading to increased computational burden when dealing with many sampling points, or is too rigid, making it challenging to capture features from irregularly shaped objects. Additionally, the number of sampling points is predetermined and cannot adjust dynamically to the shape learned by the convolutional kernel.

\subsection{Multi-scale Convolution}
Multi-scale convolution enhances the analysis of input data by utilizing convolutional kernels of different sizes, which facilitates the extraction of feature information across different scales. In comparison, standard convolution is restricted to capturing features at a single scale. Pyramidal Convolution (PyConv) \cite{PYconv} addresses this limitation by employing a hierarchical structure within each layer, utilizing a pyramid of convolutional kernels of diverse scales to process the input feature map comprehensively, as described in \cref{fig:convs}. To enhance computational efficiency and reduce the overall parameter count, the depth of each kernel—defined by the number of channels participating in the convolution operations is adaptively adjusted based on the pyramid level. Selective Kernel Networks \cite{Li2019SelectiveKN} further refine this approach by incorporating a soft attention mechanism that dynamically selects the most relevant feature maps generated from multi-scale convolutions, thereby increasing the network's adaptability to variations in spatial resolution. However, these convolutional modules still cannot adaptively adjust the sampling positions and the number of sampling points of the convolution kernel based on the sizes of various objects in the feature map.

\subsection{Motivation}
Remote sensing images exhibit considerable variety in context, with objects differing significantly in size. Using convolutional kernels of varying sizes is more effective for extracting features from different regions compared to using fixed-size kernels. Traditional shape-adaptive convolutions can modify sampling positions to align with object shapes but cannot adjust the number of sampling points based on the shape of the kernel. Additionally, some deformable strategies require learning many parameters, leading to higher computational costs. While multi-scale convolutions can capture features at various scales within the same feature map, their kernel sizes remain fixed thus cannot adaptively adjust sampling positions based on the feature map’s content. To overcome these limitations, we introduce Adaptive Rectangular Convolution (ARConv), a novel module that \textit{treats the height and width of the convolutional kernel as learnable parameters.} This allows the shape of the kernel to adjust dynamically based on the size of different objects. With sampling points evenly distributed within a rectangular deformable region, \textit{ARConv can flexibly modify sampling positions and adjust the number of points according to the average size of the learned kernels in each feature map.} Unlike conventional deformable convolutions \cite{Dai2017DeformableCN}, \textit{our approach requires learning only two parameters, minimizing computational overhead as the number of sampling points increases.} To further enhance adaptability, we apply an affine transformation to the kernel’s output, improving spatial flexibility.

\begin{figure*}[t]
  \centering
    \includegraphics[width=\linewidth, trim=0 30 0 0, clip]{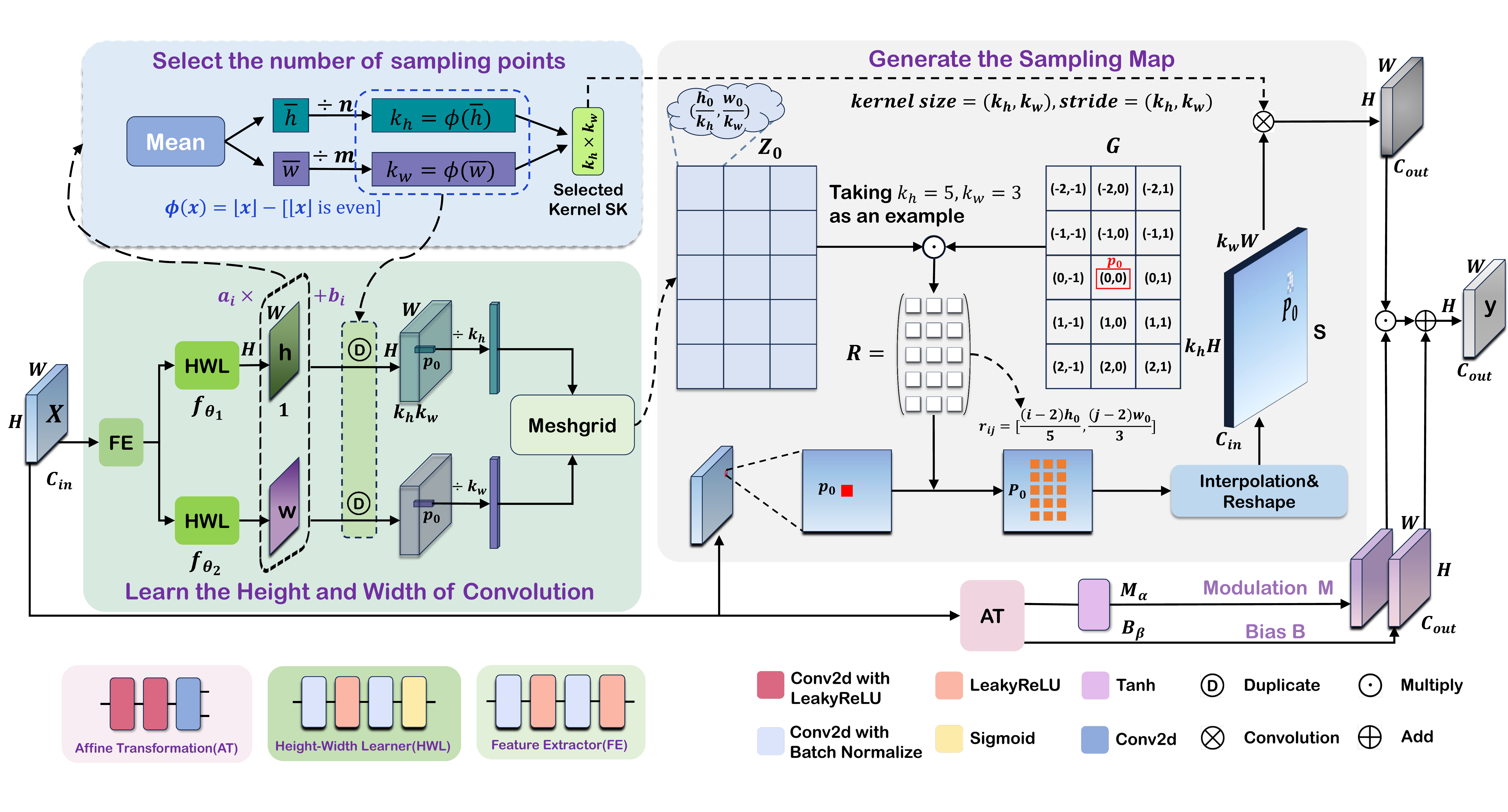}
    \caption{Overview of the ARConv architecture. This module consists of four main parts. The first part addresses the learning process of the convolution kernel's height and width. The second part focuses on the selection process for the number of sampling points of the convolution kernel. The third part simulates the generation process of the sampling map $\mathbf{S}$ using the grid center position $\mathbf{p}_0$ as an example. The final part describes the convolution operation process of ARConv.}
  \vspace{-0.5cm}
  \label{fig:ARConv}
\end{figure*}

%% file: sec/3_methods.tex
\section{Methods}
This section details the design of ARConv and ARNet. The implementation of ARConv follows four steps: (1) Learning the height and width feature maps of the convolutional kernel. (2) Selecting the number of convolutional kernel sampling points. (3) Generating the sampling map. (4) The implementation of the convolution. In ARNet, the standard
convolution layers of U-Net \cite{Unet1, Wang2021SSconvES} are replaced with ARConv modules to more effectively capture rich scale information for the pansharpening task. The overall architecture of ARConv is presented in \cref{fig:ARConv}.
\subsection{Adaptive Rectangular Convolution}
\noindent\subsubsection{Learning the Height and Width of Convolution} 

The learning process can be mathematically formulated as:
\begin{equation}
    \mathbf{y}_i = f_{\mathbf{\theta}_i}(\mathbf{X}), \quad i \in \{1,2\},
\end{equation}
where $\mathbf{X}\in\mathbb{R}^{H\times W\times C_{in}}$ represents the input feature map. $H$ and $W$ denote the height and width of the feature map, respectively, and $C_{in}$ denotes the number of input channels. Besides, $f_{\mathbf{\theta}_i}(\cdot)$ corresponds to two subnets responsible for predicting the height and width of the convolutional kernel, each consists of two components: a shared feature extractor and distinct height-width learners with $\mathbf{\theta}_i$ representing the associated parameters. The output feature maps are denoted as $\mathbf{y}_i \in \mathbb{R}^{H \times W \times 1}$, where $\mathbf{y}_1$ is the height feature map and $\mathbf{y}_2$ is the width feature map, which are referred to as $\mathbf{h}$ and $\mathbf{w}$ in the \cref{fig:ARConv}, respectively. The final layer of the height-width learner is a $\text{Sigmoid}$ function, where $\text{Sigmoid}(x)=\frac{1}{1+e^{-x}}$. So, $\mathbf{y}_i \in (0, 1)$, which only represent relative magnitudes thus cannot directly correspond to the height and width of the convolutional kernel, we adopt the following method to constrain their range of values.
\begin{align}
\mathbf{y}_i&=a_i\cdot \mathbf{y}_i + b_i, 
\quad i \in \{1,2\},
\end{align}
where $a_i$ and $b_i$ are modulation factors that constrain the range of height and width. Thus, the height of the convolutional kernel is constrained within the range $(b_1, a_1 + b_1)$, the width within $(b_2, a_2 + b_2)$.

The feature maps of height and width are fed into the second part, which will be detailed later, where the number of sampling points for the convolution kernel, $k_h \cdot k_w$, is selected. Each feature map of height and width is then replicated $k_h \cdot k_w$ times. Subsequently, a meshgrid operation is applied to generate the scaling matrix $Z_{ij} \in \mathbb{R} ^ {k_h \times k_w}$ for the convolution kernel shape at each pixel location $(i,j)$.

\subsubsection{Selecting the Number of Sampling Points} 
First, we calculate the mean of all the values in $\mathbf{y}_1$ and $\mathbf{y}_2$ to obtain the average level of the learned height and width. Then, the number of sampling points in the convolutional kernel in vertical and horizontal directions are derived from $k_h=\phi(\lfloor \frac{\mathbf{\Bar{y}}_1}{n} \rfloor),k_w=\phi(\lfloor \frac{\mathbf{\Bar{y}}_2}{m} \rfloor)$, where $\lfloor x \rfloor$ represents the floor of $x$, $m$ and $n$ denote the modulation coefficients that map the height and width of the convolutional kernel to the number of sampling points, The function $\phi(\cdot)$ can be expressed as follows:
\begin{equation}
    \phi(x) = x - [x\text{ is even}],
    \label{phi}
    \vspace{-0.1cm}
\end{equation}
Here, $[\cdot]$ denotes the Iverson bracket. Given a fixed height and width of the convolutional kernel, the larger the values of $m$ and $n$, the fewer the sampling points and the sparser their distribution. From \cref{phi}, we only select convolutional kernels with an odd number of sampling points. When $\lfloor \frac{\mathbf{\Bar{y}}_1}{n} \rfloor$ or $\lfloor \frac{\mathbf{\Bar{y}}_2}{m} \rfloor$ is even, we choose the nearest odd number that is smaller than the even number. Finally, the number of sampling points is:
\begin{equation}
    N=k_h\cdot k_w.
    \label{number}
      \vspace{-0.3cm}
\end{equation}

\begin{figure*}[t]
  \centering
    \includegraphics[width=\linewidth, trim=0 0 0 0, clip]{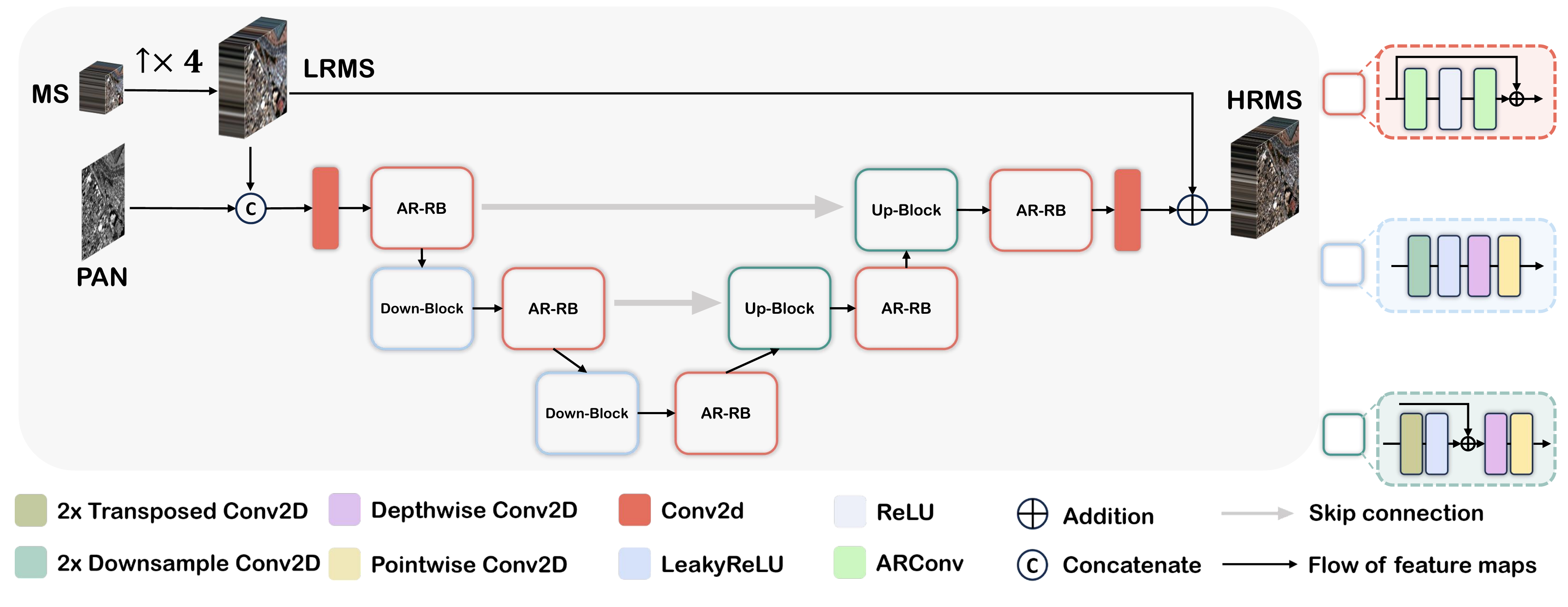}
  \caption{Overall architecture of ARNet. ARNet replaces the standard convolution in U-Net's Resblock with ARConv to create AR-Resblock. The model has down-sampling blocks to extract high-level features and up-sampling blocks to restore spatial resolution with transposed convolutions. Skip connections help transfer detailed spatial information.}
  \vspace{-0.5cm}
  \label{fig:ARNet}
\end{figure*}

\subsubsection{Generating the Sampling Map} In standard convolution, the process involves sampling from the input feature map $\mathbf{X}$ using a regular grid $\mathbf{G}$, followed by a weighted summation of these sampled values with weights $\mathbf{w}$. For example,
\begin{equation}
\mathbf{G} = \{(-1, -1), (-1, 0),\cdots,(1, 0), (1, 1)\},
\end{equation}
corresponds to a kernel covering a $3 \times 3$ region on the input map with no gaps between sampled points.

Formally, the standard convolution operation for one position $\mathbf{p}_0$ can be expressed as, 
\begin{equation}
    \mathbf{y}(\mathbf{p}_0) = \sum_{\mathbf{g}_n\in \mathbf{G}}\mathbf{w}(\mathbf{g}_n)\cdot \mathbf{x}(\mathbf{p}_0+\mathbf{g}_n),
\end{equation}
where $\mathbf{y}$ is the output feature map, $\mathbf{w}$ denotes the parameters of covolutional kernel and $\mathbf{g}_n$ represents the offsets of grid $\mathbf{G}$ relative to position $\mathbf{p}_0$.

For ARConv, we use $\mathbf{G} \in \mathbb{R}^{k_h \times k_w}$ to denote the offset matrix of the standard convolution with kernel size $k_h \times k_w$, which is shared across all pixels. The element at the $i$-th row and $j$-th column of $\mathbf{G}$, denoted by $g_{ij}$, is defined as:
\begin{equation}
    g_{ij} = \left( \frac{2i - k_h - 1}{2}, \frac{2j - k_w - 1}{2} \right).
\end{equation}

Next, we define $\mathbf{Z}_0 \in \mathbb{R}^{k_h \times k_w}$ as the scale matrix at position $\mathbf{p}_0$, which is computed in the first step. The element at the $i$-th row and $j$-th column of $\mathbf{Z}_0$, denoted by $z_{ij}$, is given by:
\begin{equation}
    z_{ij} = \left( \frac{h_0}{k_h}, \frac{w_0}{k_w} \right),
\end{equation}
where $h_0$ and $w_0$ represent the height and width of the learned convolutional kernel at position $\mathbf{p}_0$, respectively. The offset matrix of ARConv at position $\mathbf{p}_0$, denoted by $\mathbf{R}$, is then computed as:
\begin{equation}
    \mathbf{R} = \mathbf{Z}_0 \odot \mathbf{G},
\end{equation}
where $\odot$ denotes element-wise multiplication. The element at the $i$-th row and $j$-th column of $\mathbf{R}$, denoted by $r_{ij}$, is given by:
\begin{equation}
    r_{ij} = \left( \frac{(2i - k_h - 1)h_0}{2k_h}, \frac{(2j - k_w - 1)w_0}{2k_w} \right).
\end{equation}

It is evident that, in most cases, the sampling points do not coincide with the center of the grid points. Therefore, interpolation is required to estimate their pixel values. In this context, we employ bilinear interpolation, and its mathematical formulation is as follows:
\begin{equation}
    \mathbf{t}(x,y)=\mathbf{w}_x^\mathrm{T}\mathbf{T}\mathbf{w}_y,
\end{equation}
where $\mathbf{t}(x,y)$ represents the pixel value at the coordinates $(x,y)$.
\begin{equation}
    \mathbf{T}=\begin{pmatrix}
    \mathbf{t}(x_0,y_0) & \mathbf{t}(x_0,y_1)\\
    \mathbf{t}(x_1,y_0) & \mathbf{t}(x_1,y_1)
    \end{pmatrix},
\end{equation}
where $(x_0,y_0),(x_0,y_1),(x_1,y_0),(x_1,y_1)$ are the coordinates of the four grid points closest to $(x,y)$.
\begin{equation}
    \mathbf{w}_x=\begin{pmatrix}
    1-w_x\\
    w_x
    \end{pmatrix},
    \mathbf{w}_y=\begin{pmatrix}
    1-w_y\\
    w_y
    \end{pmatrix},
\end{equation}
where $w_x=\frac{x-x_0}{x_1-x_0}, w_y=\frac{y-y_0}{y_1-y_0}$. they represent the normalized interpolation weights in the $x$-direction and $y$-direction, respectively. 

In summary, the convolution operation we proposed can be mathematically expressed as:
\begin{equation}
\mathbf{y}(\mathbf{p}_0) = \sum_{\mathbf{r}_n \in \mathbf{R}} \mathbf{w}(\mathbf{r}_n) \cdot \mathbf{t}(\mathbf{p}_0 + \mathbf{r}_n),
\vspace{-0.2cm}
\end{equation}
where $\mathbf{y}(\mathbf{p}_0)$ refers to the pixel value at position $\mathbf{p}_0$ in the output feature map $\mathbf{y}$, $\mathbf{w}$ denotes the parameters of covolutional kernel, $\mathbf{r}_n$ enumerates the elements in $\mathbf{R}$, $\mathbf{t}(\mathbf{p}_0 + \mathbf{r}_n)$ calculates the pixel value at position $\mathbf{p}_0 + \mathbf{r}_n$.

 Whether using standard convolution or our method, each pixel in an image corresponds to a sampling window during the convolution operation. In standard convolution, the sampling points are all located at the grid centers, and the sampling window simply slides across the image with a fixed stride. However, in ARConv, the size of the sampling window vary for each pixel, making the traditional approach unsuitable. In practice, rather than generating a unique convolutional kernel for every pixel, we use an equivalent approach. We adopt an expansion technique, extracting the values at the locations of the sampling points corresponding to the sampling window at each pixel and assembling them into a new grid $\mathbf{P}_0$, which replaces the original pixel $\mathbf{p}_0$, where $\mathbf{\mathbf{P}}_0 \in \mathbb{R}^{k_h\times k_w\times C_{in}},\  \mathbf{p}_0 \in \mathbb{R}^{1\times1\times C_{in}}.$ After completing the expansion for each pixel, we obtain the final sampling map $\mathbf{S}$, which belongs to $\mathbb{R}^{(k_hH)\times(k_wW)\times C_{in}}$. 

\begin{table*}[h]
\scriptsize
\centering
\caption{Performance benchmarking on the WV3 dataset was conducted using 20 reduced-resolution and 20 full-resolution samples. The top-performing results are highlighted in bold, while the second-best are indicated with an underline.}
\label{tab:results_wv3}
\input{tables/table1}
\vspace{-0.3cm}
\end{table*}

\subsubsection{The Implementation of the Convolution}
In this part, we apply a convolution to $\mathbf{S}$ for feature extraction, using a kernel size and stride both set to $(k_h,k_w)$. To introduce spatial adaptability, we apply an affine transformation to the output feature map. We use two sub-networks, $\mathbf{M}_\mathbf{\alpha}$ and $\mathbf{B}_\mathbf{\beta}$, to predict the matrices $\mathbf{M}$ and $\mathbf{B}$ for the affine transformation, with $\mathbf{\alpha}$ and $\mathbf{\beta}$ as the parameters of these networks. The final output feature map is given by:
\begin{equation}
    \mathbf{y}=\mathbf{SK}\otimes \mathbf{S}\odot \mathbf{M} \oplus \mathbf{B},
    \vspace{-0.1cm}
\end{equation}
where $\mathbf{y} \in \mathbb{R}^{H\times W\times C_{out}}$ is the output feature map. $\mathbf{SK} \in \mathbb{R}^{C_{in}\times k_h\times k_w\times C_{out}}$ is the parameter of the selected convolutional kernel, $\otimes$ represents the convolution operation, $\odot$ represents element-wise multiplication and $\oplus$ represents elements-wise plus.

\subsection{ARNet Architecture}
This section details the construction of ARNet which is shown at \cref{fig:ARNet}. Our network draws inspiration from the U-net architecture \cite{Unet1, Wang2021SSconvES}, a well-known model in image segmentation that uses an encoder-decoder structure with skip connections to retain spatial information. In ARNet, we replace the standard convolutional layers in ResBlock \cite{He2015DeepRL} with our ARConv. The data flow proceeds as follows: First, the MS image is upsampled to match the resolution of the PAN image, generating the LRMS image. Next, the PAN and LRMS images are concatenated along the channel dimension and input into the network. ARNet involves a series of downsampling and upsampling steps, with ARConv layers at different depths adapting to find the optimal parameters for feature extraction at various scales. Finally, the learned details are injected back into the LRMS image \cite{FusionNet, DiCNN}, refining it and producing the final output image with enhanced resolution and detail.

\begin{table*}[h]
  \centering
  \setlength{\tabcolsep}{9.5pt} 
\begin{minipage}[h]{0.48\linewidth}
    \centering
    \caption{Performance benchmarking on the QB dataset using 20 reduced-resolution samples. Best in bold; second best underlined.}
    \label{tab:results_qb_reduced}
        \scriptsize
    \input{tables/table2}
        \vspace{-0.3cm}
  \end{minipage}
  \hfill
\begin{minipage}[h]{0.48\linewidth}
    \centering
    \caption{Performance benchmarking on the GF2 dataset using 20 reduced-resolution samples. Best in bold; second best underlined.}
    \label{tab:results_gf2_reduced}
        \scriptsize
    \input{tables/table3}
        \vspace{-0.3cm}
  \end{minipage}
\end{table*}
  

%% file: tables/table1.tex
\begin{tabular}{l *{3}{c} *{3}{c}} 
\toprule
\multirow{2}{*}{Methods} 
& \multicolumn{3}{c}{Reduced-Resolution Metrics} 
& \multicolumn{3}{c}{Full-Resolution Metrics} \\
\cmidrule(lr){2-4} \cmidrule(lr){5-7} 
& SAM$\downarrow$ & ERGAS$\downarrow$ & Q8$\uparrow$ 
& $D_\lambda\downarrow$ & $D_s\downarrow$ & HQNR$\uparrow$ \\
\midrule
EXP \cite{EXP} &  5.800 $\pm$ 1.881  &  7.155 $\pm$ 1.878  &  0.627 $\pm$ 0.092 &  0.0232 $\pm$ 0.0066  &  0.0813 $\pm$ 0.0318  &  0.897 $\pm$ 0.036  \\ 
MTF-GLP-FS \cite{MTF-GLP-FS} &  5.316 $\pm$ 1.766  &  4.700 $\pm$ 1.597  &  0.833 $\pm$ 0.092 &   0.0197 $\pm$ 0.0078  &  0.0630 $\pm$ 0.0289  &  0.919 $\pm$ 0.035  \\ 
TV \cite{TV} &  5.692 $\pm$ 1.808  &  4.856 $\pm$ 1.434  &  0.795 $\pm$ 0.120 &   0.0234 $\pm$ 0.0061  &  0.0393 $\pm$ 0.0227  &  0.938 $\pm$ 0.027   \\ 
BSDS-PC \cite{BSDS-PC} &  5.429 $\pm$ 1.823  &  4.698 $\pm$ 1.617  &  0.829 $\pm$ 0.097 &   0.0625 $\pm$ 0.0235  &  0.0730 $\pm$ 0.0356  &  0.870 $\pm$ 0.053 \\ 
CVPR2019 \cite{CVPR19} &  5.207 $\pm$ 1.574  &  5.484 $\pm$ 1.505  &  0.764 $\pm$ 0.088 &   0.0297 $\pm$ 0.0059  &  0.0410 $\pm$ 0.0136  &  0.931 $\pm$ 0.0183  \\ 
LRTCFPan \cite{LRTCFPan} &  4.737 $\pm$ 1.412  &  4.315 $\pm$ 1.442  &  0.846 $\pm$ 0.091 &   0.0176 $\pm$ 0.0066  &  0.0528 $\pm$ 0.0258  &  0.931 $\pm$ 0.031  \\ 
\midrule
PNN \cite{PNN} &  3.680 $\pm$ 0.763  &  2.682 $\pm$ 0.648  &  0.893 $\pm$ 0.092 &   0.0213 $\pm$ 0.0080  &  0.0428 $\pm$ 0.0147  &  0.937 $\pm$ 0.021  \\ 
PanNet \cite{PanNet} &  3.616 $\pm$ 0.766  &  2.666 $\pm$ 0.689  &  0.891 $\pm$ 0.093 &  \underline{ 0.0165 $\pm$ 0.0074 } &  0.0470 $\pm$ 0.0213  &  0.937 $\pm$ 0.027   \\ 
DiCNN \cite{DiCNN} &  3.593 $\pm$ 0.762  &  2.673 $\pm$ 0.663  &  0.900 $\pm$ 0.087 &   0.0362 $\pm$ 0.0111  &  0.0462 $\pm$ 0.0175  &  0.920 $\pm$ 0.026  \\ 
FusionNet \cite{FusionNet} &  3.325 $\pm$ 0.698  &  2.467 $\pm$ 0.645  &  0.904 $\pm$ 0.090 &   0.0239 $\pm$ 0.0090  &  0.0364 $\pm$ 0.0137  &  0.941 $\pm$ 0.020 \\ 
DCFNet \cite{DCFNet} &  3.038 $\pm$ 0.585  &  2.165 $\pm$ 0.499  &  0.913 $\pm$ 0.087 &   0.0187 $\pm$ 0.0072  &  0.0337 $\pm$ 0.0054  &  0.948 $\pm$ 0.012  \\ 
LAGConv \cite{Jin2022LAGConvLA} &  3.104 $\pm$ 0.559  &  2.300 $\pm$ 0.613  &  0.910 $\pm$ 0.091 &   0.0368 $\pm$ 0.0148  &  0.0418 $\pm$ 0.0152  &  0.923 $\pm$ 0.025  \\ 
HMPNet \cite{HMPNet} &  3.063 $\pm$ 0.577  &  2.229 $\pm$ 0.545  &  0.916 $\pm$ 0.087 & 0.0184 $\pm$ 0.0073  &  0.0530 $\pm$ 0.0555  &  0.930 $\pm$ 0.011  \\ 
CMT \cite{CMT} &  2.994 $\pm$ 0.607  &  2.214 $\pm$ 0.516  &  0.917 $\pm$ 0.085  &  0.0207 $\pm$ 0.0082  &  0.0370 $\pm$ 0.0078  &  0.943 $\pm$ 0.014 \\ 
CANNet \cite{Duan2024ContentAdaptiveNC} &  \underline{2.930 $\pm$ 0.593}  &  \underline{2.158 $\pm$ 0.515}  &  \underline{0.920 $\pm$ 0.084}  & 0.0196 $\pm$ 0.0083 &  \underline{0.0301 $\pm$ 0.0074}  &  \underline{0.951 $\pm$ 0.013} \\ 
\midrule
\textbf{Proposed}&\textbf{2.885 $\pm$ 0.590}& \textbf{2.139 $\pm$ 0.528}& \textbf{0.921 $\pm$ 0.083}& \textbf{0.0146 $\pm$ 0.0059}  & \textbf{0.0279 $\pm$ 0.0068} & \textbf{0.958 $\pm$ 0.010} \\ 
\bottomrule
\end{tabular}

%% file: tables/table2.tex
\begin{tabularx}{\textwidth}{l c c c}
\toprule
\textbf{Methods} & \textbf{SAM$\downarrow$} & \textbf{ERGAS$\downarrow$} & \textbf{Q4$\uparrow$} \\ 
\midrule
EXP \cite{EXP}&8.435$\pm$1.925&11.819$\pm$1.905&0.584$\pm$0.075\\
TV \cite{TV}&7.565$\pm$1.535&7.781$\pm$0.699&0.820$\pm$0.090\\
MTF-GLP-FS \cite{MTF-GLP-FS}&7.793$\pm$1.816&7.374$\pm$0.724&0.835$\pm$0.088\\
BDSD-PC \cite{BSDS-PC}&8.089$\pm$1.980&7.515$\pm$0.800&0.831$\pm$0.090\\
CVPR19 \cite{CVPR19}&7.998$\pm$1.820&9.359$\pm$1.268&0.737$\pm$0.087\\
LRTCFPan \cite{LRTCFPan}&7.187$\pm$1.711&6.928$\pm$0.812&0.855$\pm$0.087\\
\midrule
PNN \cite{PNN}&5.205$\pm$0.963&4.472$\pm$0.373&0.918$\pm$0.094\\
PanNet \cite{PanNet}&5.791$\pm$1.184&5.863$\pm$0.888&0.885$\pm$0.092\\
DiCNN \cite{DiCNN}&5.380$\pm$1.027&5.135$\pm$0.488&0.904$\pm$0.094\\
FusionNet \cite{FusionNet}&4.923$\pm$0.908&4.159$\pm$0.321&0.925$\pm$0.090\\
DCFNet \cite{DCFNet}&4.512$\pm$0.773&3.809$\pm$0.336&0.934$\pm$0.087\\
LAGConv \cite{Jin2022LAGConvLA}&4.547$\pm$0.830&3.826$\pm$0.420&0.934$\pm$0.088\\
HMPNet \cite{HMPNet}&4.617$\pm$0.404&\textbf{3.404$\pm$0.478}&0.936$\pm$0.102\\
CMT \cite{CMT}&4.535$\pm$0.822&3.744$\pm$0.321&0.935$\pm$0.086\\
CANNet \cite{Duan2024ContentAdaptiveNC}&\underline{4.507$\pm$0.835}&3.652$\pm$0.327&\underline{0.937$\pm$0.083}\\
\midrule
\textbf{Proposed}&\textbf{4.430$\pm$0.811}& \underline{3.633$\pm$0.327}&\textbf{0.939$\pm$0.081}\\
\bottomrule
\end{tabularx}

%% file: tables/table3.tex
\begin{tabularx}{\textwidth}{l c c c }
\toprule
\textbf{Methods} & \textbf{SAM$\downarrow$} & \textbf{ERGAS$\downarrow$} & \textbf{Q4$\uparrow$} \\ 
\midrule
EXP \cite{EXP}&1.820$\pm$0.403&2.366$\pm$0.554&0.812$\pm$0.051\\
TV \cite{TV}&1.918$\pm$0.398&1.745$\pm$0.405&0.905$\pm$0.027\\
MTF-GLP-FS \cite{MTF-GLP-FS}&1.655$\pm$0.385&1.589$\pm$0.395&0.897$\pm$0.035\\
BDSD-PC \cite{BSDS-PC}&1.681$\pm$0.360&1.667$\pm$0.445&0.892$\pm$0.035\\
CVPR19 \cite{CVPR19}&1.598$\pm$0.353&1.877$\pm$0.448&0.886$\pm$0.028\\
LRTCFPan \cite{LRTCFPan}&1.315$\pm$0.283&1.301$\pm$0.313&0.932$\pm$0.033\\
\midrule
PNN \cite{PNN}&1.048$\pm$0.226&1.057$\pm$0.236&0.960$\pm$0.010\\
PanNet \cite{PanNet}&0.997$\pm$0.212&0.919$\pm$0.191&0.967$\pm$0.010\\
DiCNN \cite{DiCNN}&1.053$\pm$0.231&1.081$\pm$0.254&0.959$\pm$0.010\\
FusionNet \cite{FusionNet}&0.974$\pm$0.212&0.988$\pm$0.222&0.964$\pm$0.009\\
DCFNet \cite{DCFNet}&0.872$\pm$0.169&0.784$\pm$0.146&0.974$\pm$0.009\\
LAGConv \cite{Jin2022LAGConvLA}&0.786$\pm$0.148&0.687$\pm$0.113&0.981$\pm$0.008\\
HMPNet \cite{HMPNet}&0.803$\pm$0.141&\textbf{0.564$\pm$0.099}&0.981$\pm$0.020\\
CMT \cite{CMT}&0.753$\pm$0.138&0.648$\pm$0.109&0.982$\pm$0.007\\
CANNet \cite{Duan2024ContentAdaptiveNC}&\underline{0.707$\pm$0.148}&0.630$\pm$0.128&\textbf{0.983$\pm$0.006}\\
\midrule
\textbf{Proposed}&$\textbf{0.698}$$\pm$$\textbf{0.149}$&\underline{0.626$\pm$0.127}&\underline{0.983$\pm$0.007}\\
\bottomrule
\end{tabularx}

%% file: sec/4_experiments.tex
\section{Experiments}
\subsection{Datasets, Metrics and Training Details}
We evaluate the effectiveness of our method on several datasets, including 8-band data captured by the WorldView3 (WV3) sensor, as well as 4-band data captured by the QuickBird (QB) and the GaoFen-2 (GF2) sensors. Although we use a supervised learning approach, ground truth data is not directly available, so we apply Wald’s protocol \cite{Deng2021DetailID, Wald1997FusionOS} to construct our dataset. All three datasets are accessible from a public repository \cite{Pancollection}. For test sets with different resolutions, we use different evaluation metrics. Specifically, we employ SAM \cite{Boardman1993AutomatingSU}, ERGAS \cite{Wald2002DataFD}, and Q8 \cite{Garzelli2009HypercomplexQA} to assess the performance of ARNet on the reduced-resolution dataset, and $D_s$, $D_\lambda$, and HQNR \cite{Arienzo2022FullResolutionQA} to evaluate its performance on the full-resolution dataset. During training, we employ the $l_1$ loss function along with the Adam optimizer \cite{Kingma2014AdamAM}, using a batch size of 16. Given that our method involves selecting convolutional kernels based on the learned height and width—an approach that can complicate convergence—we designate the initial 100 epochs as an exploratory phase. During this phase, we allow the model to explore different configurations. After these 100 epochs, we randomly select a combination of convolutional kernels from the 16 batches based on the results obtained and then fix this selection for the remainder of the training. \textit{Further details on the dataset and training procedure are provided in the supplementary material \cref{datasets} and \ref{train}}.

\begin{figure*}[t]
  \centering
\includegraphics[width=\linewidth, trim=0 10 0 0, clip]{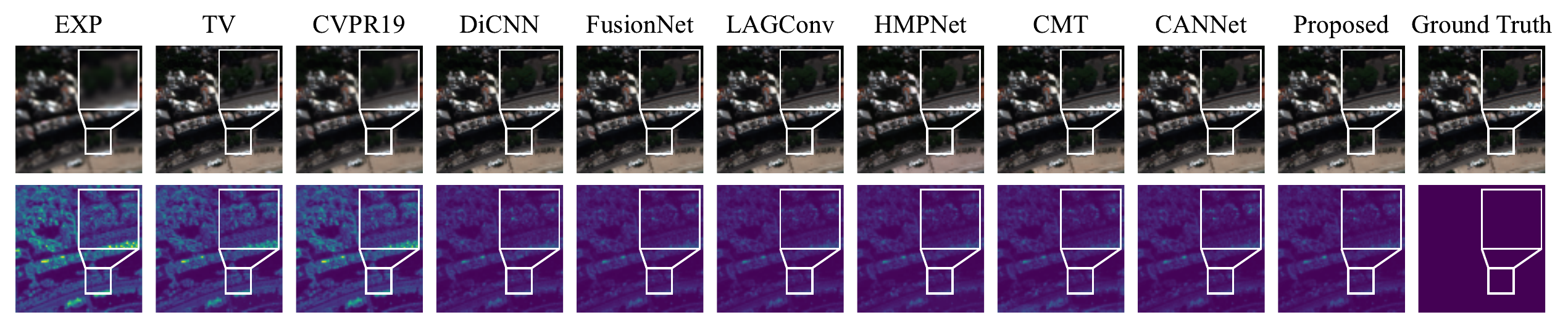}
  \caption{Qualitative comparison of benchmark methods on WV3 reduced-resolution dataset. Top: RGB outputs; Bottom: residuals vs. ground truth. \textit{See Suppl. \cref{MR} for details.}}
  \label{fig:gt}
 \vspace{-0.4cm}
\end{figure*}

\subsection{Results}
The outstanding performance of ARNet has been thoroughly demonstrated through a comprehensive evaluation on the WV3, QB, and GF2 benchmark datasets. \cref{tab:results_wv3} to \ref{tab:results_gf2_reduced} provide a detailed comparison of ARNet against various state-of-the-art techniques, including traditional methods, general deep learning methods, and specialized convolution-based deep learning approaches similar to proposed work, such as LAGConv \cite{Jin2022LAGConvLA} and CANConv \cite{Duan2024ContentAdaptiveNC}, \textit{more details can be found in supplementary material \cref{benchmark methods}.} The results clearly indicate that ARNet consistently delivers high-quality performance across different datasets, showcasing remarkable robustness. Moreover, visual assessments reveal that the images generated by ARNet are the closest to the ground truth, illustrating the ability of our convolutional approach to effectively adapt to varying object sizes and extract features at appropriate scales. \textit{For further details on the benchmark tests and visual examples, please refer to the supplementary material \cref{MR}.}

\begin{table}[h]
    \centering
 \setlength{\tabcolsep}{12pt} 
 \begin{minipage}[h]{1.0\linewidth}
    \centering
    \caption{Ablation study on WV3 reduced-res dataset: HWA (height and width adaptation), NSPA (sampling points adaptation), AT (affine transformation).}
    \label{tab:6}
        \scriptsize
    \input{tables/table6}
    \vspace{-0.3cm}
  \end{minipage}
\end{table}

\subsection{Ablation Study}
To assess the impact of different components in ARConv, we conducted ablation experiments by selectively removing certain modules: (a) without height and width adaptation, (b) without the number of sampling points adaptation, and (c) without affine transformation. The results are shown in \cref{tab:6}. The performance drop in (a) and (b) highlights the effectiveness of ARConv in adapting to different object sizes. In (c), the sharp decline indicates limited flexibility in our deformation strategy, which is effectively mitigated by introducing spatial adaptability through affine transformation. Notably, the computational cost of this transformation does not increase with kernel size.

\vspace{-0.1cm}
\begin{figure*}[t]
  \centering
\includegraphics[width=\linewidth, trim=0 0 0 15, clip]{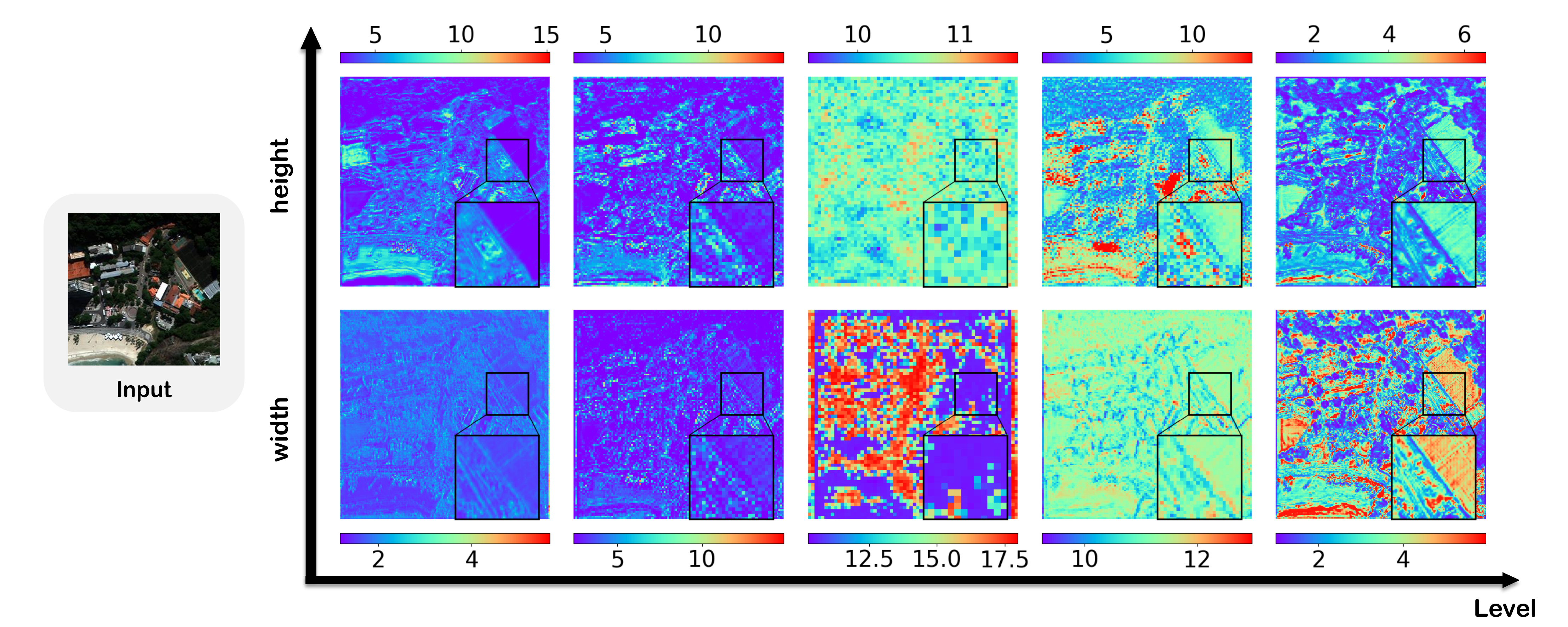}
  \caption{Heatmaps of the heights and widths learned at each pixel by convolutional kernels at different layers. The input image is a sample from the WV3 dataset. In the heatmaps, various colors represent different heights and widths captured by the convolutional kernels.}
      \vspace{-0.4cm}
  \label{fig:HEAT}
\end{figure*}

\begin{table}[h]
\begin{minipage}[h]{1.0\linewidth}
 \setlength{\tabcolsep}{14pt} 
    \centering
    \caption{Performance of different convolution kernel height and width learning ranges on WV3 reduced-resolution dataset.}
    \label{tab:5}
        \scriptsize
    \input{tables/table5}
    \vspace{-0.5cm}
  \end{minipage}
\end{table}

\subsection{Discussion}
\noindent\textbf{Different Height and Width Learning Ranges:}
To assess the impact of different convolutional kernel heights and widths on ARNet's performance, we designed five sets of experiments with varying height and width ranges: (a) 1-3, (b) 1-9, (c) 1-18, (d) 1-36, and (e) 1-63. In (a), the kernel size is fixed at $3 \times 3$, while in (b) to (e), the maximum kernel size is $7 \times 7$. As shown in \cref{tab:5}, ARNet’s performance initially improves with an increasing height and width range but declines beyond an optimal setting in case (c). This pattern arises because a smaller range results in densely packed sampling points that capture excessive noise, while a larger range spreads sampling points too sparsely, reducing the kernel's ability to capture fine details.
\begin{table}[h]
  \centering
  \setlength{\tabcolsep}{11pt} 
  \begin{minipage}[h]{1.0\linewidth}
    \centering
    \caption{Performance on the WV3 reduced-resolution dataset when replacing convolution kernels in other pansharpening methods with ARConv.}
    \label{tab:4}
        \scriptsize
    \input{tables/table4}
    \vspace{-0.3cm}
  \end{minipage}
\end{table}

\noindent\textbf{Replacing the Convolutional Modules in other Networks: }
We integrate ARConv as a plug-and-play module, replacing the original convolution layers in pansharpening networks such as FusionNet \cite{FusionNet}, LAGNet \cite{Jin2022LAGConvLA}, and CANNet \cite{Duan2024ContentAdaptiveNC} to demonstrate ARConv's effectiveness. The results in \cref{tab:4} indicate that ARConv significantly enhances the performance of these networks. \textit{Additional details on this experiment are provided in the supplementary materials \cref{RCE}.}

\noindent\textbf{Convolutional Kernel Visualization: }\cref{fig:HEAT} shows the height and width feature maps learned by the convolutional kernels at different layers of ARNet. The overall heatmaps reveal the contours of various objects in the RGB image, especially in the outermost layers of the network. Although the intermediate layers appear disordered, they capture deeper semantic information, such as object sizes in the RGB image. For example, in the height heatmap of the fourth layer, the outline of a tilted building is faintly visible, with a thin blue line along the edges. This indicates that the learned heights of the convolutional kernels are smaller at the edges, reflecting the adaptation of the kernels to the building's dimensions. \textit{Please refer to supplementary material in \cref{MV} for more visualizations.}
\begin{table}[h]
  \centering
  \setlength{\tabcolsep}{12.2pt} \begin{minipage}[h]{1.0\linewidth}
    \centering
    \caption{Performance comparison between ARConv and DCNv2 on WV3 reduced-resolution.}
    \label{tab:7}
        \scriptsize
    \input{tables/table7}
        \vspace{-0.4cm}
  \end{minipage}
\end{table}

\noindent\textbf{Comparison with DCNv2: }We removed the affine transformation from ARConv and adopted the same modulation method as DCNv2 \cite{Zhu2018DeformableCV}. Both models were trained for 600 epochs on the WV3 dataset. The results are shown in \cref{tab:7}, where it is evident that our performance surpasses DCNv2. This may be because the deformation strategy in DCNv2 requires learning a larger number of parameters, which can hinder convergence in sharpening tasks.

%% file: tables/table6.tex
\begin{tabularx}{\textwidth}{l c c c}
\toprule
\textbf{Methods} & \textbf{SAM$\downarrow$} & \textbf{ERGAS$\downarrow$} & \textbf{Q8$\uparrow$} \\ 
\midrule
\textbf{(a) No HWA}&2.925$\pm$0.593& 2.171$\pm$0.557& 0.920$\pm$0.085\\
\textbf{(b) No NSPA}&2.911$\pm$0.603& 2.152$\pm$0.565& 0.921$\pm$0.083\\
\textbf{(c) No AT}&3.020$\pm$0.614& 2.269$\pm$0.562& 0.916$\pm$0.085\\
\textbf{Proposed}&2.885$\pm$0.590& 2.139$\pm$0.528& 0.921$\pm$0.083\\
\bottomrule
\end{tabularx}

%% file: tables/table5.tex
\begin{tabularx}{\textwidth}{l c c c}
\toprule
\textbf{Methods} & \textbf{SAM$\downarrow$} & \textbf{ERGAS$\downarrow$} & \textbf{Q8$\uparrow$} \\ 
\midrule
\textbf{(a) 1-3}&2.923$\pm$0.600& 2.164$\pm$0.546& 0.919$\pm$0.085\\
\textbf{(b) 1-9}&2.896$\pm$0.588&2.145$\pm$0.544&0.921$\pm$0.084\\
\textbf{(c) 1-18}&2.885$\pm$0.590&2.139$\pm$0.528&0.921$\pm$0.083\\
\textbf{(d) 1-36}&3.044$\pm$0.646&2.216$\pm$0.578& 0.916$\pm$0.087\\
\textbf{(e) 1-63}&3.066$\pm$0.593& 2.249$\pm$0.554& 0.912$\pm$0.095\\
\bottomrule
\end{tabularx}

%% file: tables/table4.tex
\begin{tabularx}{\textwidth}{l c c c}
\toprule
\textbf{Methods} & \textbf{SAM$\downarrow$} & \textbf{ERGAS$\downarrow$} & \textbf{Q8$\uparrow$}\\ 
\midrule
\textbf{FusionNet} \cite{FusionNet} & 3.325$\pm$0.698 & 2.467$\pm$0.645 & 0.904$\pm$0.090\\
\textbf{AR-FusionNet}&3.171$\pm$0.650&2.395$\pm$0.630&0.911$\pm$0.087\\
\midrule
\textbf{LAGNet} \cite{Jin2022LAGConvLA}&3.104$\pm$0.559&2.300$\pm$0.613&0.910$\pm$0.091\\
\textbf{AR-LAGNet}&3.083$\pm$0.643&2.277$\pm$0.547&0.916$\pm$0.085\\
\midrule
\textbf{CANNet} \cite{Duan2024ContentAdaptiveNC}&2.930$\pm$0.593&2.158$\pm$0.515&0.920$\pm$0.084\\
\textbf{AR-CANNet}&2.885$\pm$0.590&2.139$\pm$0.528&0.921$\pm$0.083\\
\bottomrule
\end{tabularx}

%% file: tables/table7.tex
\begin{tabularx}{\textwidth}{l c c c}
\toprule
\textbf{Methods} & \textbf{SAM$\downarrow$} & \textbf{ERGAS$\downarrow$} & \textbf{Q8$\uparrow$} \\ 
\midrule
\textbf{Ours}&2.881$\pm$0.590& 2.149$\pm$0.531& 0.921$\pm$0.084\\
\textbf{DCNv2} \cite{Zhu2018DeformableCV}&3.151$\pm$0.679& 2.425$\pm$0.656& 0.915$\pm$0.083\\
\bottomrule
\end{tabularx}

%% file: sec/5_conclusion.tex
\section{Conclusion}
In conclusion, we introduce an adaptive rectangular convolution module, ARConv, which dynamically learns height- and width-adaptive convolution kernels for each pixel based on the varying sizes of objects in the input image. By adjusting the number of sampling points according to the learned scale, ARConv overcomes the traditional limitations of fixed sampling shapes and point counts in convolution kernels. Integrated seamlessly into U-net as a plug-and-play module, ARConv forms ARNet, which has demonstrated outstanding performance across multiple datasets. Additionally, the visualization studies confirm that our convolutional kernels can effectively adjust their height and width based on the size and shape of objects, offering a novel solution for the pansharpening task.

%% file: sec/X_suppl.tex
\clearpage
\setcounter{page}{1}
\maketitlesupplementary

\begin{abstract}
The supplementary material provides a detailed description of the experimental setup using the ARConv module, covering several key aspects. It includes an overview of the dataset composition, followed by the configuration of the training process. Additionally, the material offers a brief introduction to the benchmark methods and outlines the specifics of the convolution kernel replacement experiment. Finally, it presents further result comparisons and visualizations to support the findings.
\end{abstract}
\vspace{-0.5cm}

\section{Details on Experiments}
\subsection{Datasets}
\label{datasets}
The experimental data used in this study is captured by three different sensors: WorldView3 (WV3), QuickBird (QB), and Gao-Fen2 (GF2). A downsampling process is used to simulate and build our dataset, which includes three training sets corresponding to the three sensors. Each training set is paired with both reduced-resolution and full-resolution test sets, enabling comprehensive model evaluation across different image qualities. The training sets consist of PAN/LRMS/GT image pairs, with dimensions of $64\times64$, $64\times64\times C$, and $64\times64\times C$, respectively. The WV3 training set contains 9,714 PAN/LRMS/GT image pairs (C = 8), the QB training set contains 17,139 pairs (C = 4), and the GF2 training set contains 19,809 pairs (C = 4). The corresponding reduced-resolution test sets for these three training sets each consist of 20 PAN/LRMS/GT image pairs, with dimensions of $256\times256$, $256\times256\times C$, and $256\times256\times C$, respectively. The full-resolution dataset includes 20 RAN/LRMS image pairs, with dimensions of $512\times512$, $512\times512\times C$. These datasets are publicly available through the PanCollection repository \cite{Pancollection}.

\subsection{Training Details}
\label{train}
This section provides a detailed description of the training details for all our experiments, focusing on aspects such as the loss function, optimizer, batch size, number of training epochs, exploratory phase epochs, convolution kernel height and width learning range, initial learning rate, and learning rate decay methods. In all experiments, the loss function used is $l_1$loss, the optimizer is Adam optimizer \cite{Kingma2014AdamAM}, the batch size is 16, the initial learning rate is 0.0006, the learning rate decays by a factor of 0.8 every 200 epochs and the exploratory phase consists of 100 epochs. The purpose of the exploratory phase is to address the challenge of convergence when selecting the number of convolution kernel sampling points based on the average learned height and width of the kernels. After the exploratory phase, we randomly select a set of convolution kernel sampling point combinations and keep them fixed during the subsequent training process. The remaining configuration differences are shown in the \cref{tab:details_on_training}.

\subsection{Benchmark Methods}
\label{benchmark methods}
In the main text, we provide a detailed comparison between the proposed method and several established approaches. To facilitate this comparison, \cref{tab:benchmark_methods} presents a concise overview of the benchmark methods used in our study. The table is divided into two parts by a horizontal line, with traditional methods listed above the line and deep learning methods below the line.

\subsection{Replacing Convolution Experiment}
\label{RCE}
In FusionNet, the original architecture consists of four standard residual blocks. In AR-FusionNet, we replace the convolution layers in the two middle residual blocks with our proposed ARConv. This modification results in a total of four ARConv layers in the network, which enhances its ability to capture more complex features. Similarly, in LAGNet, which has five standard residual blocks, we replace the convolution layers in the second and fourth blocks with ARConv. This strategic placement allows us to evaluate ARConv in a deeper network structure, providing a comparison with other models. ARNet and CANNet are constructed similarly, with each replacing the standard convolution modules in the U-Net architecture \cite{Unet1, Wang2021SSconvES} with their respective proposed convolution modules. Specifically, in CANNet, all standard convolutions are replaced with ARConv, thus transforming it into ARNet. This provides a natural comparison between the two networks, offering valuable insights into the impact of the different convolution techniques. The training set for all three experiments is WV3, other training details can be found in \cref{tab:RCE}.

\begin{table*}[htbp]
    \centering
 \setlength{\tabcolsep}{7.5pt} 
 \begin{minipage}[h]{1.0\linewidth}
    \centering
    \caption{The different configurations for replacing convolution experiment. The first three columns represent the experiment name, the number of training epochs, and the convolution kernel height and width learning range. The subsequent columns, Layer1-10, represent the final number of sampling points for each of the ten convolution layers.}
    \label{tab:RCE}
        \scriptsize
    \input{tables/table10}
  \end{minipage}
\end{table*}
\hfill
\begin{table*}[htbp]
    \centering
 \setlength{\tabcolsep}{7.5pt} 
 \begin{minipage}[h]{1.0\linewidth}
    \centering
    \caption{The different configurations for all experiments except replacing convolution experiment which is detailed in \cref{RCE}. The first three columns represent the experiment name, the number of training epochs, and the convolution kernel height and width learning range, where "HWR" stands for Height and Width Range. The subsequent columns, Layer1-10, represent the final number of sampling points for each of the ten convolution layers in ARNet. The names of the first three experiments correspond to their respective training datasets, while all subsequent experiments use the WV3 dataset for training.}
    \label{tab:details_on_training}
        \scriptsize
    \input{tables/table8}
  \end{minipage}
\end{table*}
\hfill
\begin{table}[h]
    \centering
 \setlength{\tabcolsep}{7.5pt} 
 \begin{minipage}[h]{1.0\linewidth}
    \centering
    \caption{Performance benchmarking on the QB dataset using 20 full-resolution samples. Best in bold; second best underlined.}
    \label{tab:QB_FR}
        \scriptsize
    \input{tables/table11}
  \end{minipage}
  \vspace{-0.5cm}
\end{table}
\begin{table}[h]
    \centering
 \setlength{\tabcolsep}{7.5pt} 
 \begin{minipage}[h]{1.0\linewidth}
    \centering
    \caption{Performance benchmarking on the GF2 dataset using 20 full-resolution samples. Best in bold; second best underlined.}
    \label{tab:GF2_FR}
        \scriptsize
    \input{tables/table12}
      \vspace{-0.5cm}
  \end{minipage}
\end{table}
\hfill
\begin{table*}[htbp]
    \centering
 \setlength{\tabcolsep}{7.5pt} 
 \begin{minipage}[h]{1.0\linewidth}
    \centering
    \caption{A brief introduction to various benchmark methods.}
    \label{tab:benchmark_methods}
        \scriptsize
    \input{tables/table9}
  \end{minipage}
\end{table*}
\subsection{More Results}
\label{MR}
\begin{figure*}[t]
  \centering
\includegraphics[width=\linewidth, trim=0 0 0 0, clip]{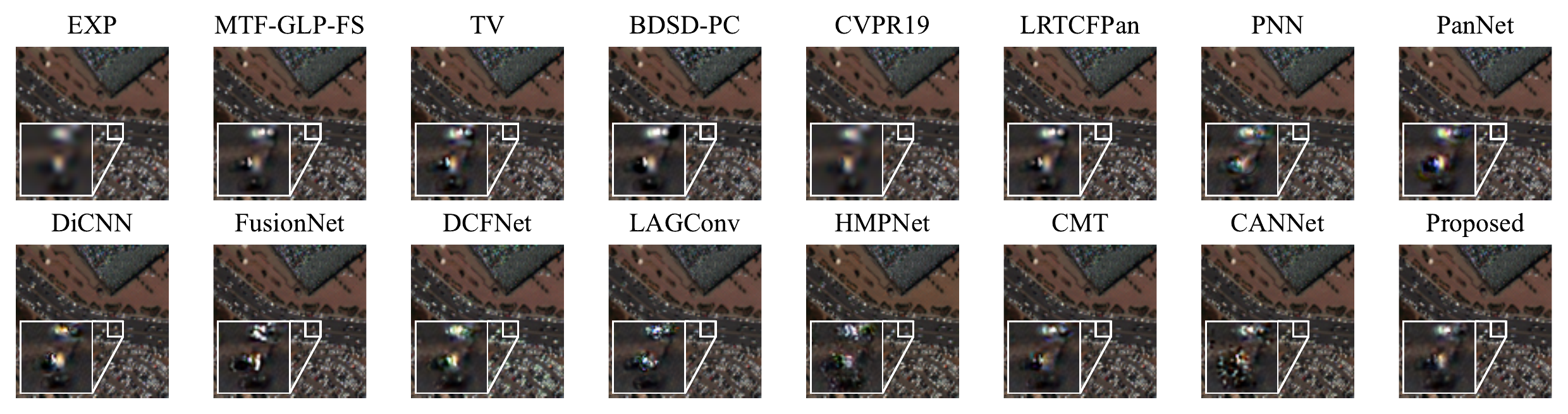}
  \caption{Comparison of qualitative results among benchmark methods on WV3 full-resolution dataset. The first row displays the RGB outputs, and the second row shows the residual relative to the ground truth. \textbf{Zoom in for best view.}}
  \label{fig:gt_wv3_fr_18}
\end{figure*}

\begin{figure*}[t]
  \centering
\includegraphics[width=\linewidth, trim=0 0 0 0, clip]{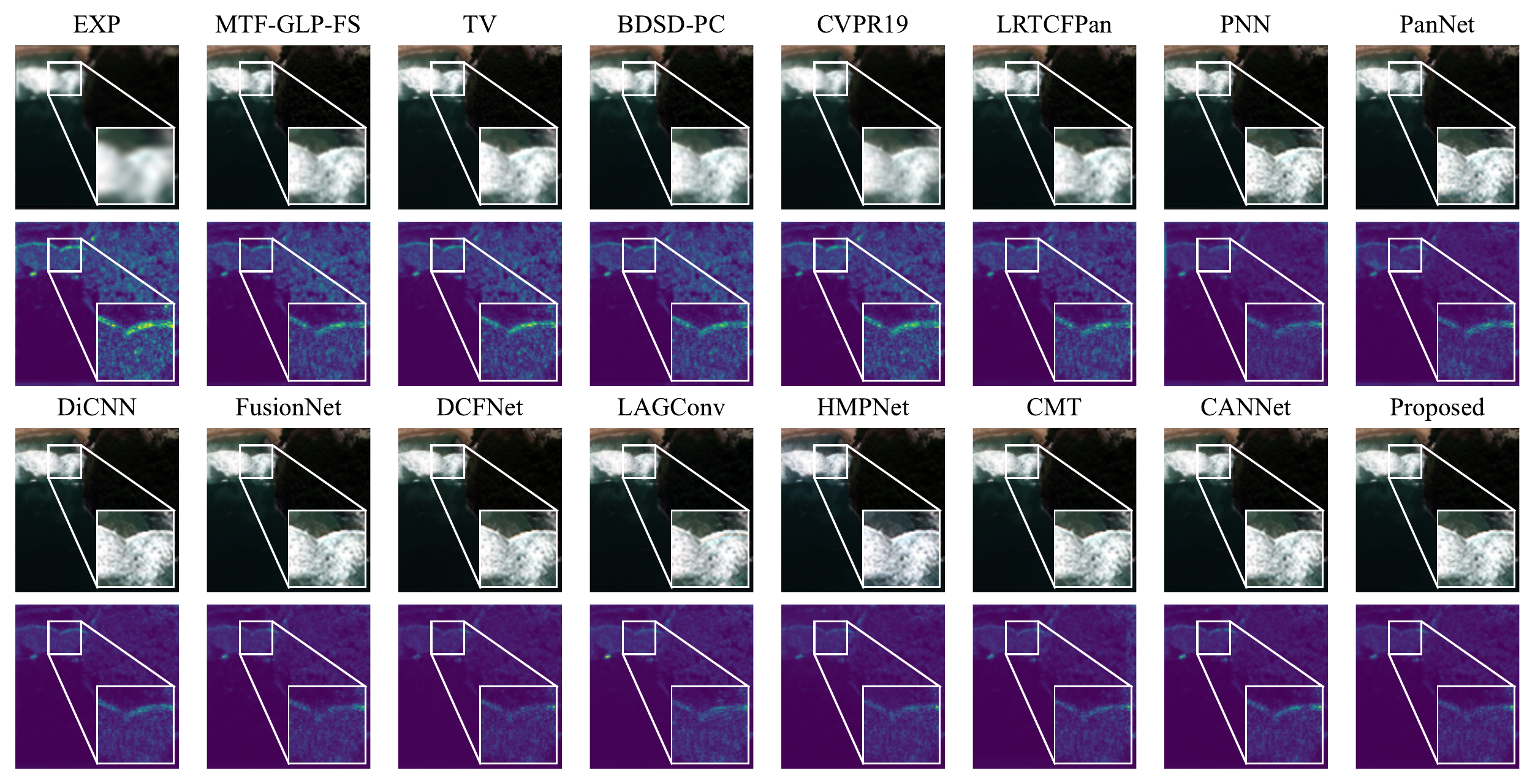}
  \caption{Comparison of qualitative results among benchmark methods on WV3 reduced-resolution dataset. The first row displays the RGB outputs, and the second row shows the residual relative to the ground truth. \textbf{Zoom in for best view.}}
  \label{fig:gt_wv3_rr_13}
\end{figure*}

\begin{figure*}[t]
  \centering
\includegraphics[width=\linewidth, trim=0 0 0 0, clip]{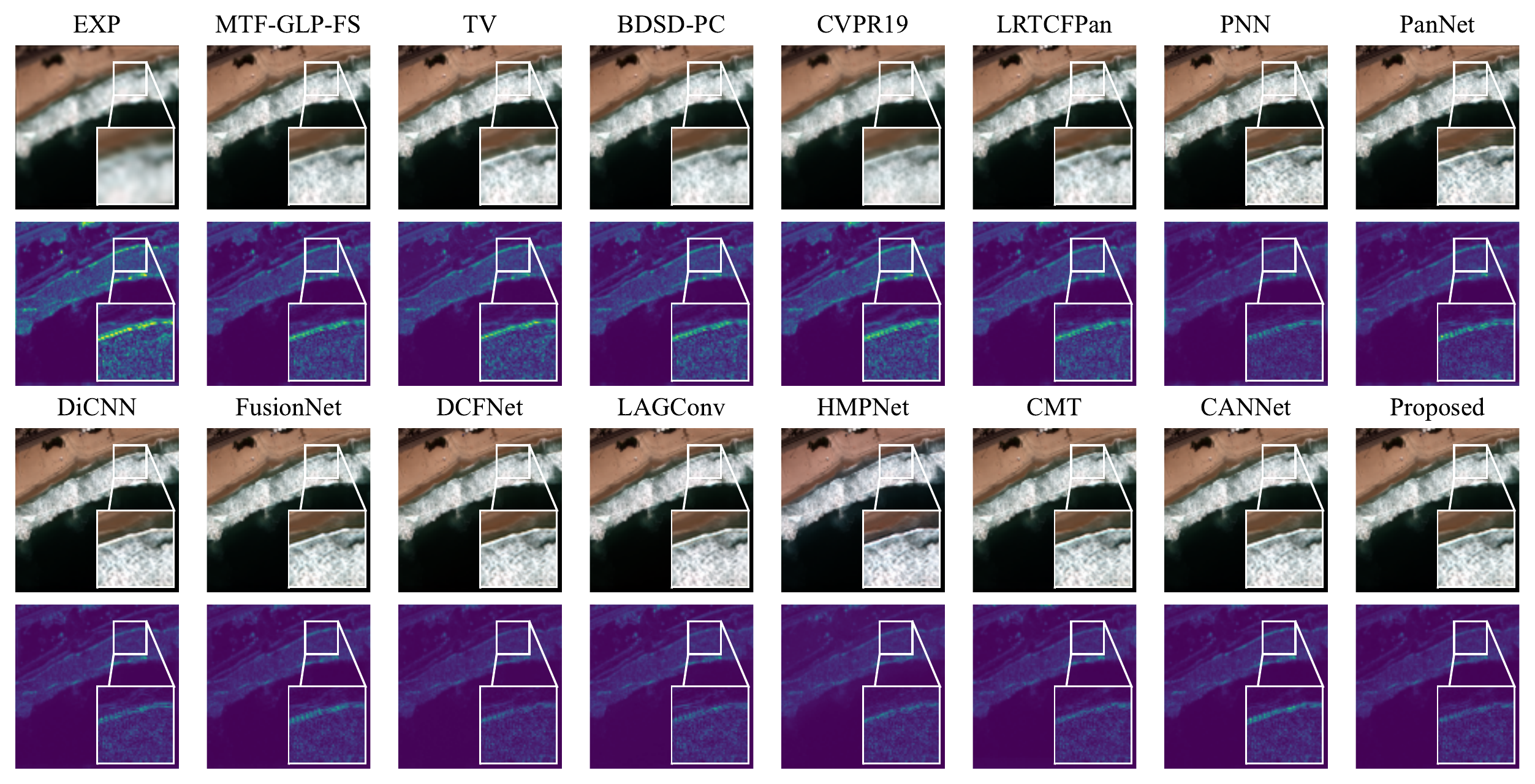}
  \caption{Comparison of qualitative results among benchmark methods on WV3 reduced-resolution dataset. The first row displays the RGB outputs, and the second row shows the residual relative to the ground truth. \textbf{Zoom in for best view.}}
  \label{fig:gt_wv3_rr_12}
\end{figure*}

\begin{figure*}[t]
  \centering
\includegraphics[width=\linewidth, trim=0 0 0 0, clip]{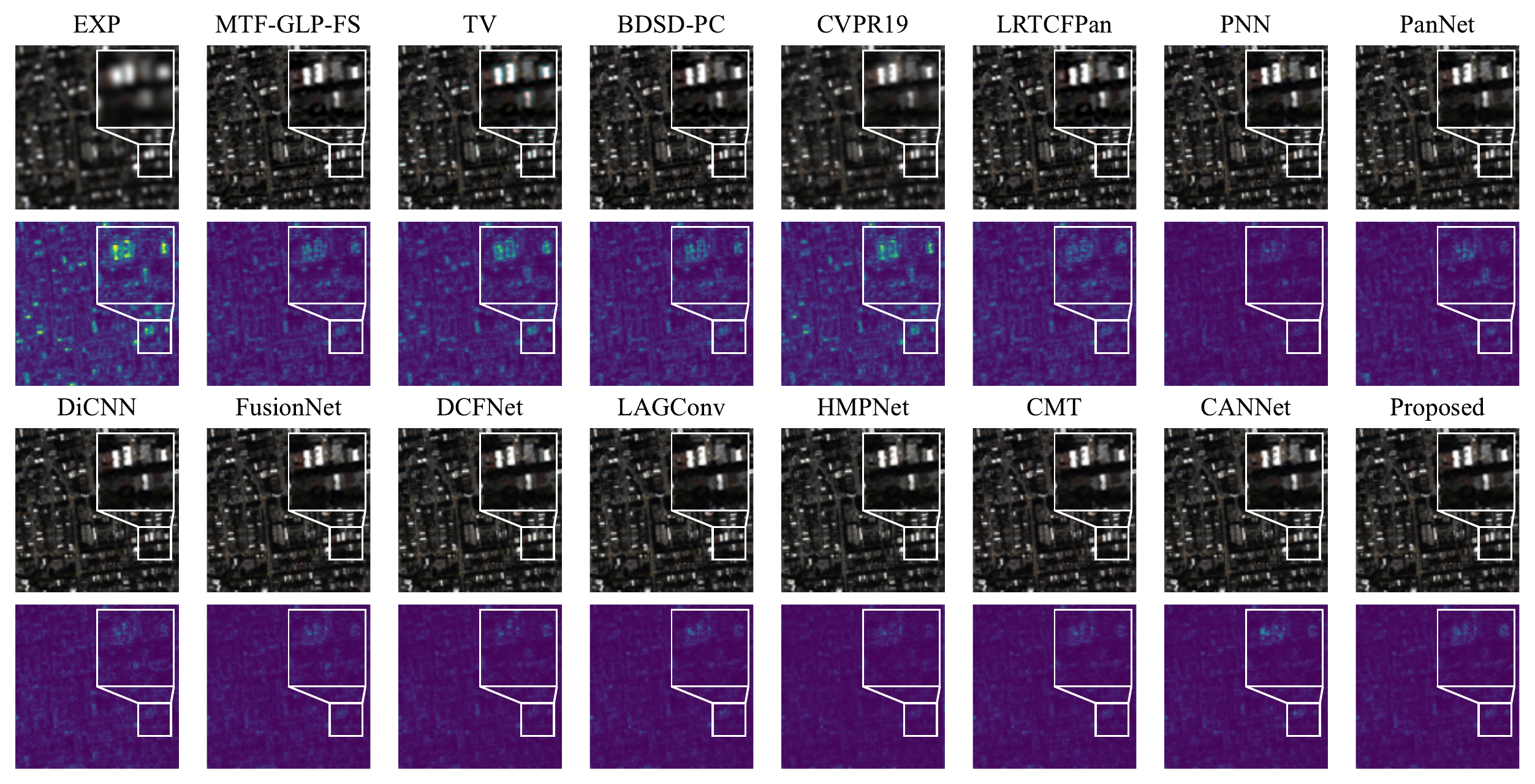}
  \caption{Comparison of qualitative results among benchmark methods on QB reduced-resolution dataset. The first row displays the RGB outputs, and the second row shows the residual relative to the ground truth. \textbf{Zoom in for best view.}}
  \label{fig:gt_qb_rr_8}
\end{figure*}

\begin{figure*}[t]
  \centering
\includegraphics[width=\linewidth, trim=0 0 0 0, clip]{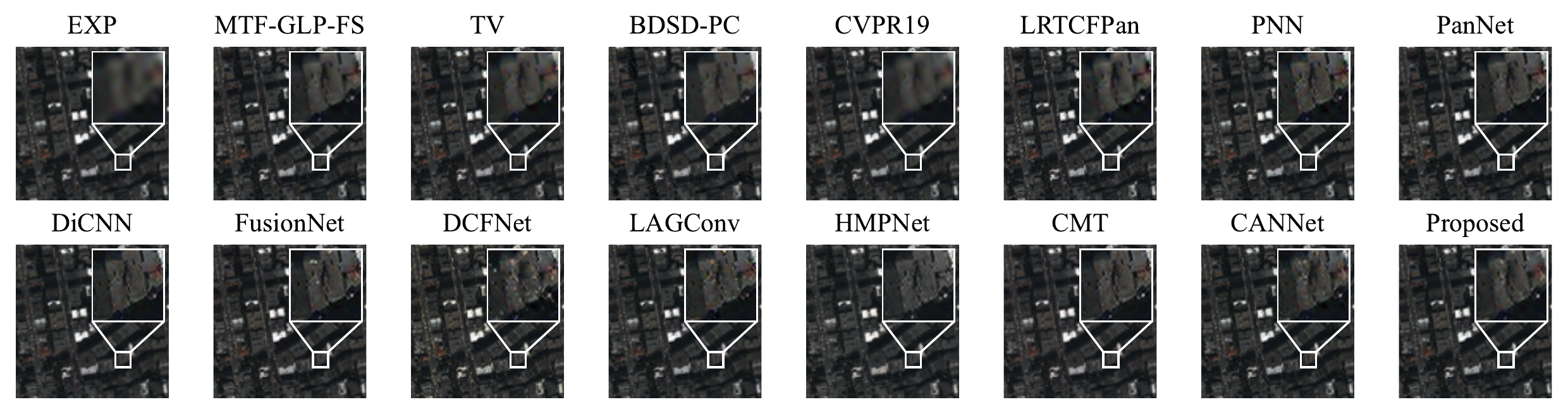}
  \caption{Comparison of qualitative results among benchmark methods on QB full-resolution dataset. The first row displays the RGB outputs, and the second row shows the residual relative to the ground truth. \textbf{Zoom in for best view.}}
  \label{fig:gt_qb_fr_6}
\end{figure*}

\begin{figure*}[t]
  \centering
\includegraphics[width=\linewidth, trim=0 0 0 0, clip]{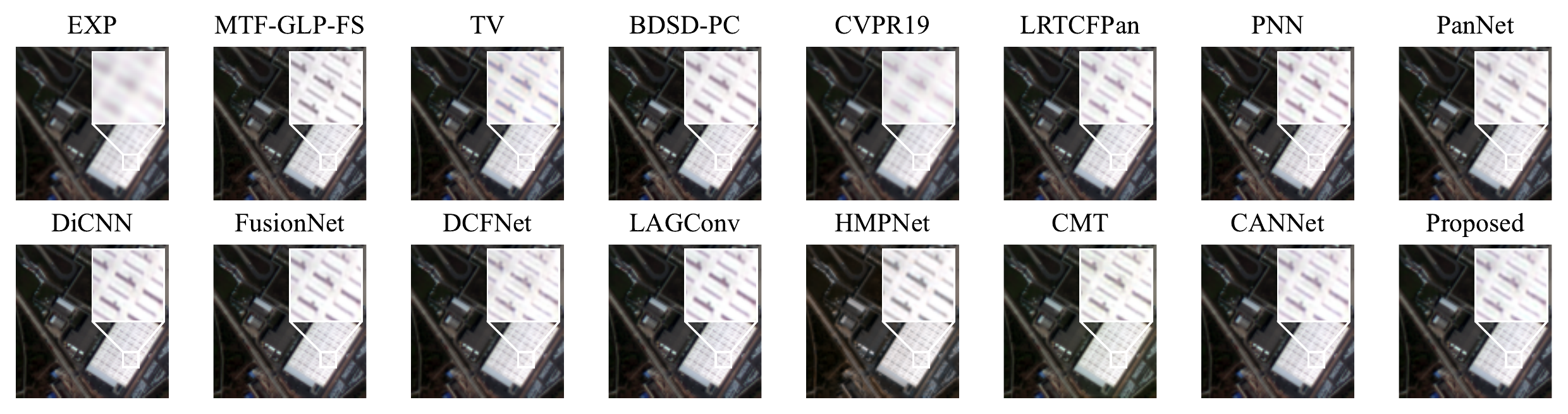}
  \caption{Comparison of qualitative results among benchmark methods on GF2 full-resolution dataset. The first row displays the RGB outputs, and the second row shows the residual relative to the ground truth. \textbf{Zoom in for best view.}}
  \label{fig:gt_gf2_fr_0}
\end{figure*}

\begin{figure*}[t]
  \centering
\includegraphics[width=\linewidth, trim=0 0 0 0, clip]{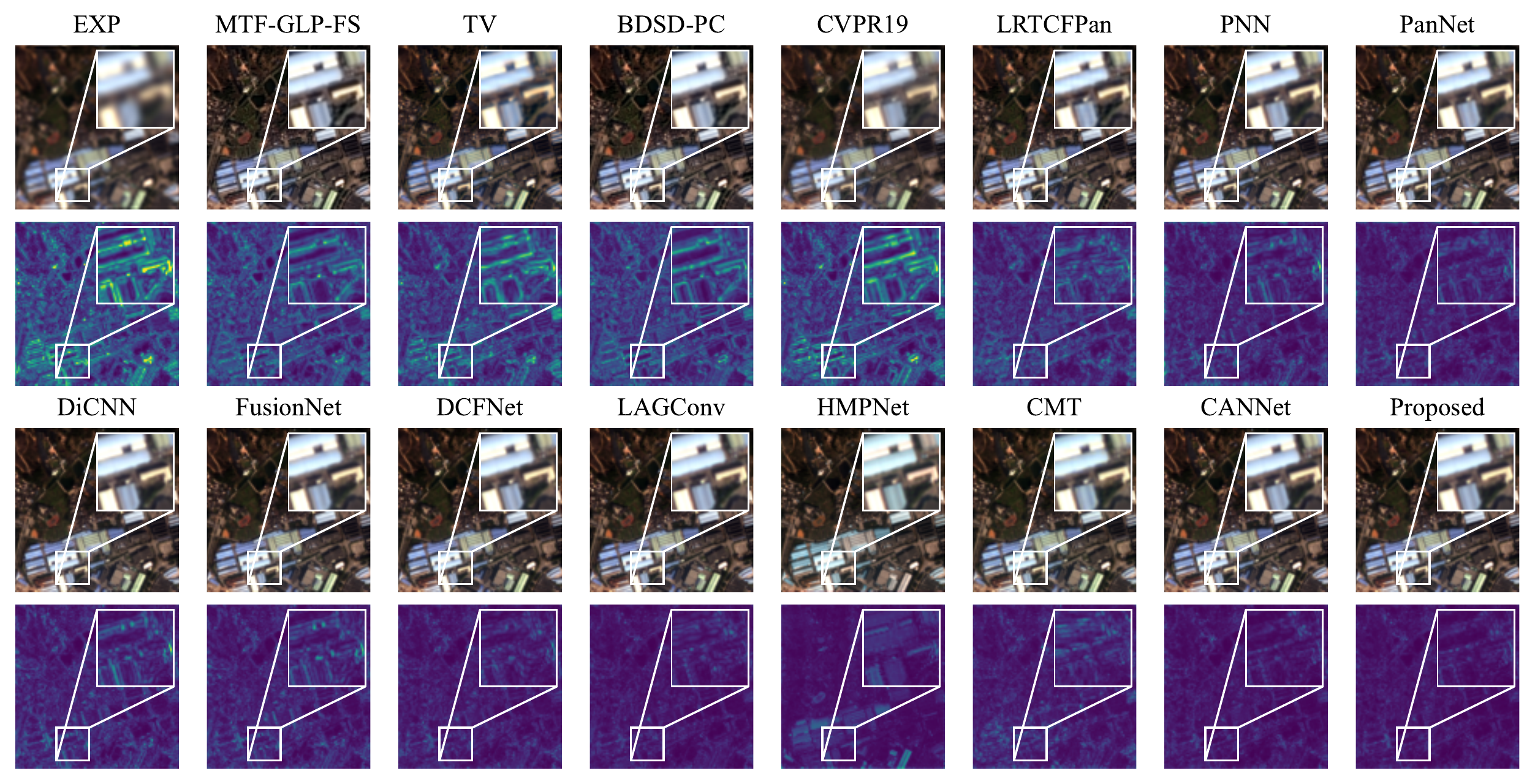}
  \caption{Comparison of qualitative results among benchmark methods on GF2 reduced-resolution dataset. The first row displays the RGB outputs, and the second row shows the residual relative to the ground truth. \textbf{Zoom in for best view.}}
  \label{fig:gt_gf2_rr_8}
\end{figure*}

\begin{figure*}[t]
  \centering
\includegraphics[width=\linewidth, trim=0 0 0 0, clip]{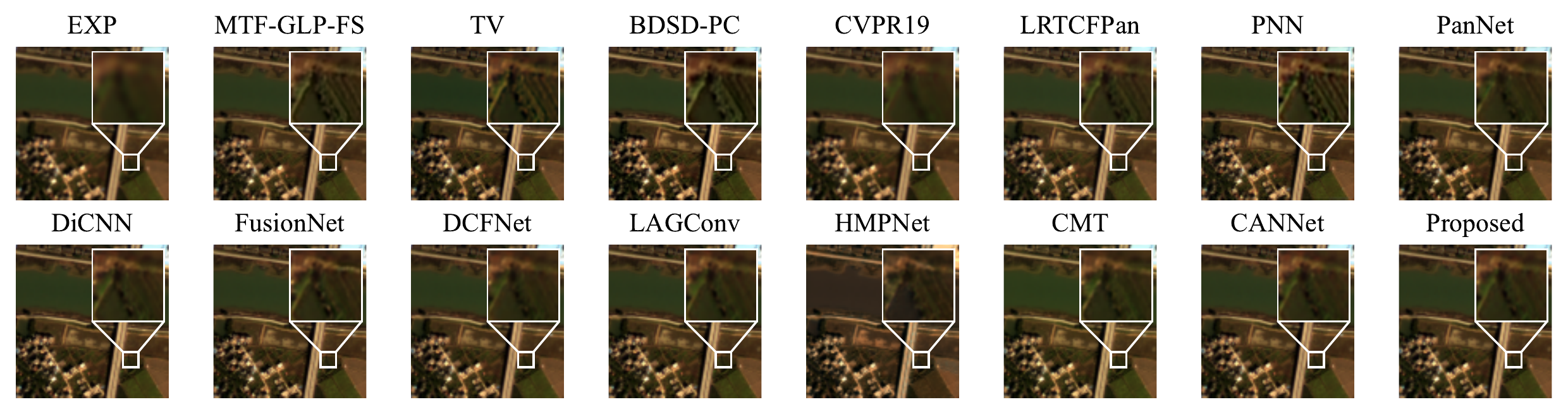}
  \caption{Comparison of qualitative results among benchmark methods on GF2 full-resolution dataset. The first row displays the RGB outputs, and the second row shows the residual relative to the ground truth. \textbf{Zoom in for best view.}}
  \label{fig:gt_gf2_fr_18}
\end{figure*}

\cref{tab:QB_FR} and \ref{tab:GF2_FR} present the performance benchmarks on the full-resolution QB and GF2 datasets, evaluating the effectiveness of various methods. Among the three metrics, $D_\lambda$ measures the network's ability to capture spectral information, while $D_s$ reflects the network’s capacity to preserve spatial details. The metric $\text{HQNR} = (1-D_\lambda)(1-D_s)$ provides a comprehensive evaluation of the network’s overall performance and is considered the most critical metric for assessing methods on full-resolution datasets. \cref{fig:gt_wv3_fr_18} to \ref{fig:gt_gf2_fr_18} display the outputs of ARNet compared to various benchmark methods on both reduced-resolution and full-resolution test sets from the WV3, QB, and GF2 datasets. Additionally, residual maps between the outputs and the ground truth are provided for the reduced-resolution datasets. These figures and tables strongly demonstrate the robustness of our proposed method across multiple datasets.
\subsection{More Visualizations}
\label{MV}
\begin{figure*}[t]
  \centering
\includegraphics[width=\linewidth, trim=0 0 0 0, clip]{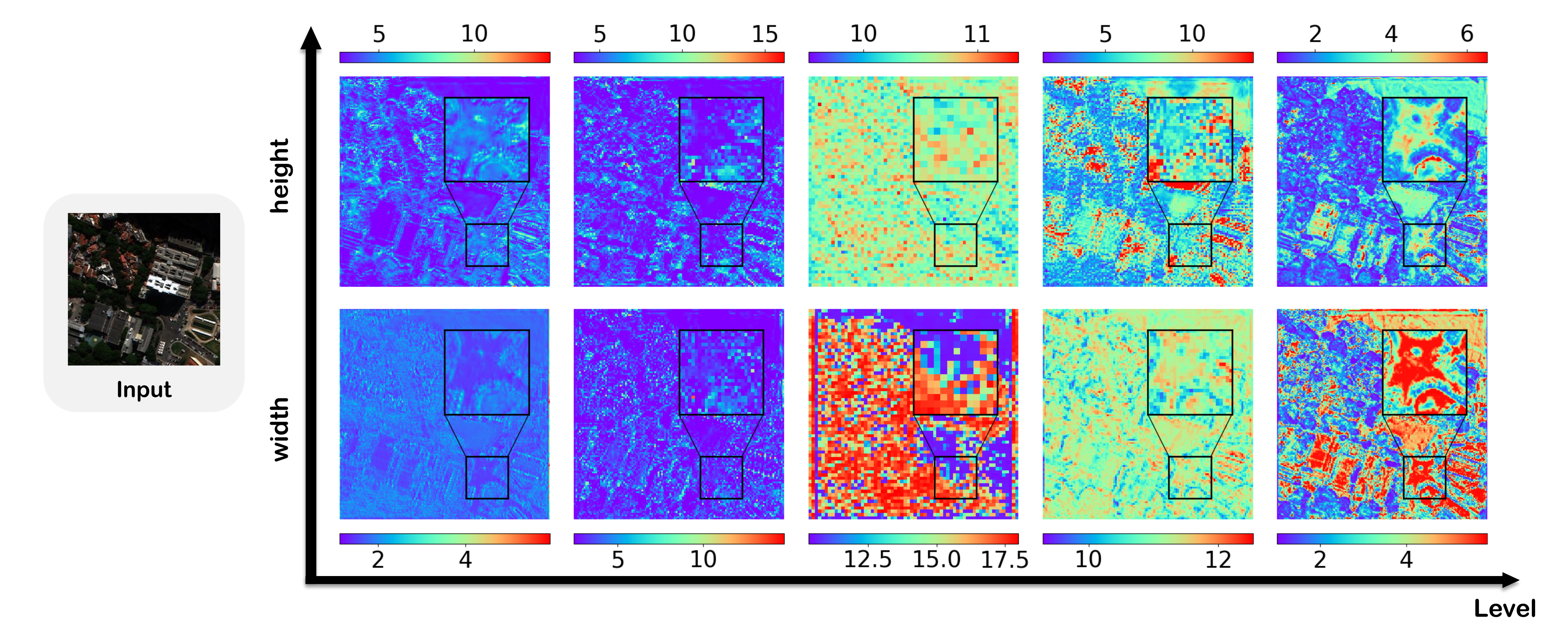}
  \caption{Heatmaps of the heights and widths learned at each pixel by convolutional kernels at different layers. The input image is a sample from the WV3 dataset.}
  \label{fig:heatmap_wv3_5}
\end{figure*}

\begin{figure*}[t]
  \centering
\includegraphics[width=\linewidth, trim=0 0 0 0, clip]{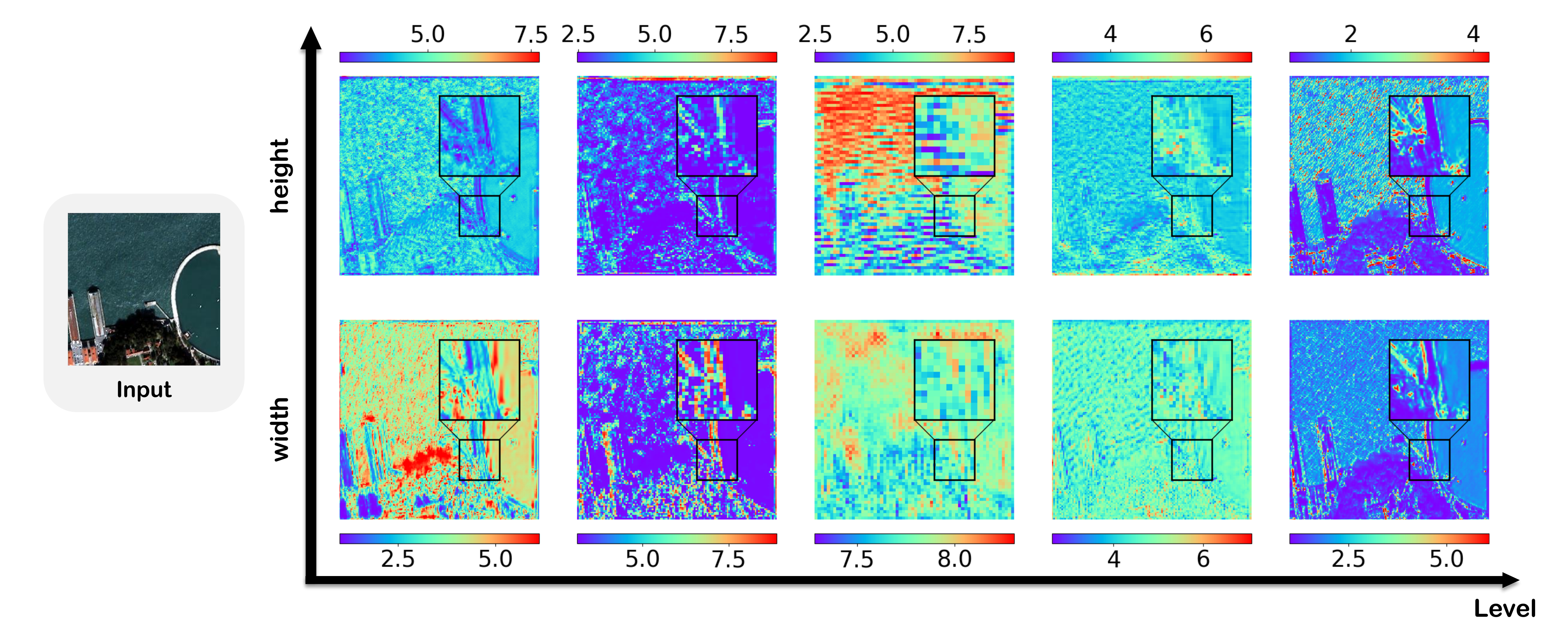}
  \caption{Heatmaps of the heights and widths learned at each pixel by convolutional kernels at different layers. The input image is a sample from the QB dataset.}
  \label{fig:heatmap_qb_2}
\end{figure*}

\begin{figure*}[t]
  \centering
\includegraphics[width=\linewidth, trim=0 0 0 0, clip]{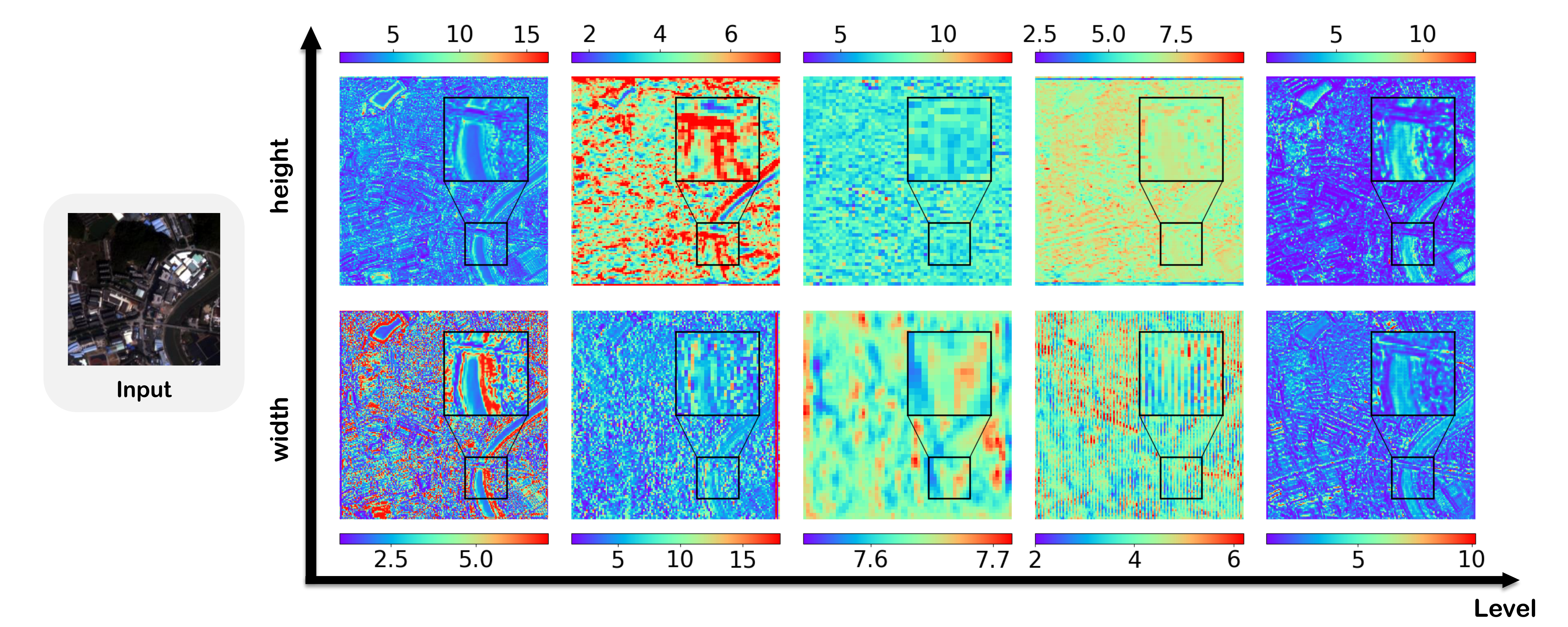}
  \caption{Heatmaps of the heights and widths learned at each pixel by convolutional kernels at different layers. The input image is a sample from the GF2 dataset.}
  \label{fig:heatmap_gf2_10}
\end{figure*}

\begin{figure*}[t]
  \centering
\includegraphics[width=\linewidth, trim=0 0 0 0, clip]{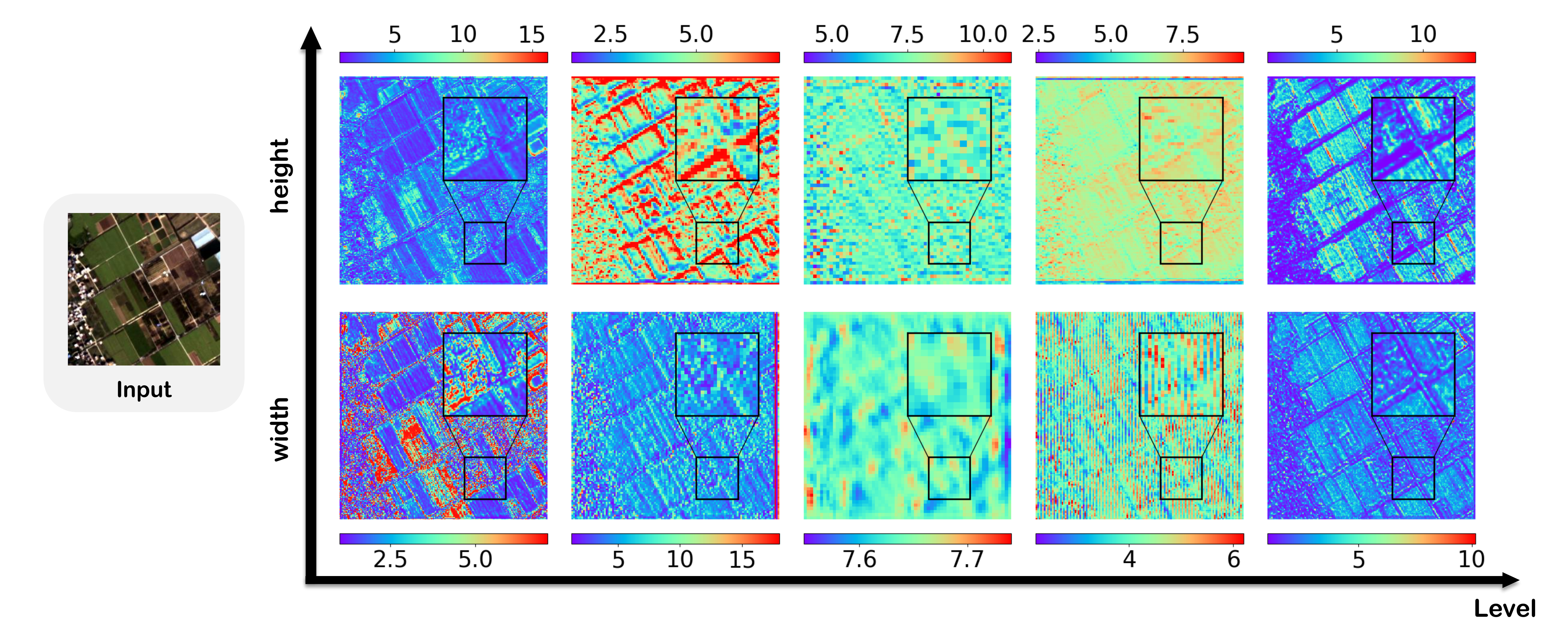}
  \caption{Heatmaps of the heights and widths learned at each pixel by convolutional kernels at different layers. The input image is a sample from the GF2 dataset.}
  \label{fig:heatmap_gf2_20}
\end{figure*}

In ARNet, there are a total of five AR-Resblocks, each containing two ARConv layers. For our analysis, we select one ARConv from each block and visualize the heatmaps corresponding to the height and width of its learned convolution kernel. The heatmaps, shown in \cref{fig:heatmap_wv3_5} to \ref{fig:heatmap_gf2_20}, provide valuable insight into the relationship between the kernel shapes and the object sizes present in the feature maps. This adaptability highlights the flexibility of our approach in handling various object scales and offers compelling evidence of the effectiveness of our method in dynamically adjusting to different input characteristics, making it a powerful tool for tasks that require precise and scalable feature extraction.

%% file: tables/table10.tex
\begin{tabular}{lclccccc}
\toprule
\textbf{Experiment} & \textbf{Epochs} & \textbf{Range} & \textbf{Layer1-2} & \textbf{Layer3-4} & \textbf{Layer5-6} & \textbf{Layer7-8} & \textbf{Layer9-10} \\
\midrule
\textbf{AR-FusionNet} & $400$ & $1-9$ & $3 \times 5, 7 \times 7$ & $7 \times 3, 5 \times 5$ & $-$ & $-$ & $-$ \\
\textbf{AR-LAGNet} & $220$ & $1-9$ & $5 \times 3, 5 \times 3$ &$ 5 \times 3, 7 \times 7 $& $-$ & $-$ & $-$ \\
\textbf{AR-CANNet} & $600$ & $1-18$ & $3\times3,3\times3$ & $7\times5,3\times5$ & $3\times3,3\times3$ & $3\times3,5\times5$& $3\times5,3\times3$\\
\bottomrule
\end{tabular}

%% file: tables/table8.tex
\begin{tabular}{lclccccc}
\toprule
\textbf{Experiment} & \textbf{Epochs} & \textbf{Range} & \textbf{Layer1-2} & \textbf{Layer3-4} & \textbf{Layer5-6} & \textbf{Layer7-8} & \textbf{Layer9-10} \\
\midrule
\textbf{WV3} &$600$ & $1-18$ & $3\times3,3\times3$ & $7\times5,3\times5$ & $3\times3,3\times3$ & $3\times3,5\times5$& $3\times5,3\times3$\\
\textbf{QB} & $200$ & $1-9$ & $3\times3,3\times5$ & $5\times7,3\times3$ & $5\times3,3\times3$ & $3\times3,7\times7$& $3\times3,3\times3$\\
\textbf{GF2} & $630$ & $1-18$ & $3\times3,3\times3$ & $3\times7,3\times5$ & $3\times3,3\times3$ & $3\times3,5\times3$& $3\times3,3\times3$\\
\midrule
\textbf{Ablation study (a)} & $600$ & $ 3-3$ & $3\times3,3\times3$ & $3\times3,3\times3$ & $3\times3,3\times3$ & $3\times3,3\times3$& $3\times3,3\times3$\\
\textbf{Ablation study (b)} & $600$ & $1-18$ & $3\times3,3\times3$ & $3\times3,3\times3$ & $3\times3,3\times3$ & $3\times3,3\times3$& $3\times3,3\times3$\\
\textbf{Ablation study (c)} & $600$ & $1-18$ & $3\times3,3\times3$ & $5\times5,3\times3$ & $3\times3,3\times3$ & $3\times3,3\times3$& $3\times3,3\times3$\\
\midrule
\textbf{HWR}$1-3$ & $600$ & $1-3$ & $3\times3,3\times3$ & $3\times3,3\times3$ & $3\times3,3\times3$ & $3\times3,3\times3$& $3\times3,3\times3$\\
\textbf{HWR}$1-9$ & $600$ & $1-9$ & $5\times3,3\times3$ & $3\times3,3\times3$ & $5\times3,5\times3$ & $3\times3,3\times3$& $5\times3,3\times3$\\
\textbf{HWR}$1-18$ & $600$ & $1-18$ & $3\times3,3\times3$ & $7\times5,3\times5$ & $3\times3,3\times3$ & $3\times3,5\times5$& $3\times5,3\times3$\\
\textbf{HWR}$1-36$ & $600$ & $1-36$ & $3\times3,3\times3$ & $3\times3,3\times3$ & $3\times3,5\times5$ & $5\times3,3\times3$& $3\times3,3\times3$\\
\textbf{HWR}$1-63$ & $600$ & $1-63$ & $3\times3,3\times3$ & $5\times5,5\times5$ & $5\times5,5\times5$ & $3\times5,5\times5$& $3\times3,3\times3$\\
\midrule
\textbf{Comparison with DCNv2} & $600$ & $1-18$ & $3\times3,3\times3$ & $3\times3,3\times3$ & $3\times3,3\times3$ & $3\times3,3\times3$& $3\times3,3\times3$\\
\bottomrule
\end{tabular}

%% file: tables/table11.tex
\begin{tabularx}{\textwidth}{l c c c}
\toprule
\textbf{Methods} & \textbf{$D_\lambda\downarrow$} & \textbf{$D_s\downarrow$} & \textbf{HQNR$\uparrow$} \\ 
\midrule
EXP \cite{EXP}&0.0436$\pm$0.0089&0.1502$\pm$0.0167&0.813$\pm$0.020\\
TV \cite{TV}&0.0465$\pm$0.0146&0.1500$\pm$0.0238&0.811$\pm$0.034\\
MTF-GLP-FS \cite{MTF-GLP-FS}&0.0550$\pm$0.0142&0.1009$\pm$0.0265&0.850$\pm$0.037\\
BDSD-PC \cite{BSDS-PC}&0.1975$\pm$0.0334&0.1636$\pm$0.0483&0.672$\pm$0.058\\
CVPR19 \cite{CVPR19}&0.0498$\pm$0.0119&0.0783$\pm$0.0170&0.876$\pm$0.023\\
LRTCFPan \cite{LRTCFPan}&\textbf{0.0226$\pm$0.0117}&0.0705$\pm$0.0351&0.909$\pm$0.044\\
\midrule
PNN \cite{PNN}&0.0577$\pm$0.0110&0.0624$\pm$0.0239&0.844$\pm$0.030\\
PanNet \cite{PanNet}&0.0426$\pm$0.0112&0.1137$\pm$0.0323&0.849$\pm$0.039\\
DiCNN \cite{DiCNN}&0.0947$\pm$0.0145&0.1067$\pm$0.0210&0.809$\pm$0.031\\
FusionNet \cite{FusionNet}&0.0572$\pm$0.0182&0.0522$\pm$0.0088&0.894$\pm$0.021\\
DCFNet \cite{DCFNet}&0.0469$\pm$0.0150&0.1239$\pm$0.0269&0.835$\pm$0.016\\
LAGConv \cite{Jin2022LAGConvLA}&0.0859$\pm$0.0237&0.0676$\pm$0.0136&0.852$\pm$0.018\\
HMPNet \cite{HMPNet}&0.1832$\pm$0.0542&0.0793$\pm$0.0245&0.753$\pm$0.065\\
CMT \cite{CMT}&0.0504$\pm$0.0122&\textbf{0.0368$\pm$0.0075}&0.915$\pm$0.016\\
CANNet \cite{Duan2024ContentAdaptiveNC}&\underline{0.0370$\pm$0.0129}&0.0499$\pm$0.0092&\underline{0.915$\pm$0.012}\\
\midrule
\textbf{Proposed}&0.0384$\pm$0.0148& \underline{0.0396$\pm$0.0090}&\textbf{0.924$\pm$0.0191}\\
\bottomrule
\end{tabularx}

%% file: tables/table12.tex
\begin{tabularx}{\textwidth}{l c c c}
\toprule
\textbf{Methods} & \textbf{$D_\lambda\downarrow$} & \textbf{$D_s\downarrow$} & \textbf{HQNR$\uparrow$} \\ 
\midrule
EXP \cite{EXP}&\underline{0.0180$\pm$0.0081}&0.0957$\pm$0.0209&0.888$\pm$0.023\\
TV \cite{TV}&0.0346$\pm$0.0137&0.1429$\pm$0.0282&0.828$\pm$0.035\\
MTF-GLP-FS \cite{MTF-GLP-FS}&0.0553$\pm$0.0430&0.1118$\pm$0.0226&0.839$\pm$0.044\\
BDSD-PC \cite{BSDS-PC}&0.0759$\pm$0.0301&0.1548$\pm$0.0280&0.781$\pm$0.041\\
CVPR19 \cite{CVPR19}&0.0307$\pm$0.0127&0.0622$\pm$0.0101&0.909$\pm$0.017\\
LRTCFPan \cite{LRTCFPan}&0.0325$\pm$0.0269&0.0896$\pm$0.0141&0.881$\pm$0.023\\
\midrule
PNN \cite{PNN}&0.0317$\pm$0.0286&0.0943$\pm$0.0224&0.877$\pm$0.036\\
PanNet \cite{PanNet}&\textbf{0.0179$\pm$0.0110}&0.0799$\pm$0.0178&0.904$\pm$0.020\\
DiCNN \cite{DiCNN}&0.0369$\pm$0.0132&0.0992$\pm$0.0131&0.868$\pm$0.016\\
FusionNet \cite{FusionNet}&0.0350$\pm$0.0124&0.1013$\pm$0.0134&0.867$\pm$0.018\\
DCFNet \cite{DCFNet}&0.0240$\pm$0.0115&0.0659$\pm$0.0096&0.912$\pm$0.012\\
LAGConv \cite{Jin2022LAGConvLA}&0.0284$\pm$0.0130&0.0792$\pm$0.0136&0.895$\pm$0.020\\
HMPNet \cite{HMPNet}&0.0819$\pm$0.0499&0.1146$\pm$0.0126&0.813$\pm$0.049\\
CMT \cite{CMT}&0.0225$\pm$0.0116&\textbf{0.0433$\pm$0.0096}&\textbf{0.935$\pm$0.014}\\
CANNet \cite{Duan2024ContentAdaptiveNC}&0.0194$\pm$0.0101&0.0630$\pm$0.0094&0.919$\pm$0.011\\
\midrule
\textbf{Proposed}&0.0189$\pm$0.0097&\underline{0.0515$\pm$0.0099}&\underline{0.931$\pm$0.012}\\
\bottomrule
\end{tabularx}

%% file: tables/table9.tex
\begin{tabularx}{\textwidth}{l l l}
\toprule
\textbf{Method} & \textbf{Year} & \textbf{Introduction}\\
\midrule
\textbf{EXP} \cite{EXP} & \textbf{2002} & Upsamples the MS image.\\
\textbf{MTF-GLP-FS} \cite{MTF-GLP-FS} & \textbf{2018} & Focuses on a regression-based approach for pansharpening, specifically for the estimation of injection coefficients at full resolution.\\
\textbf{TV} \cite{TV} & \textbf{2014} & Uses total variation to regularize an ill-posed problem in a widely used image formation model.\\
\textbf{BSDS-PC} \cite{BSDS-PC} & \textbf{2019} & Addresses the limitations of the traditional BDSD method when fusing multispectral images with more than four spectral bands.\\
\textbf{CVPR2019} \cite{CVPR19} & \textbf{2019} & Proposes a new variational pan-sharpening model based on local gradient constraints to improve spatial preservation.\\
\textbf{LRTCFPan} \cite{LRTCFPan} & \textbf{2023} & Proposes a novel low-rank tensor completion (LRTC)-based framework for multispectral pansharpening.\\
\midrule
\textbf{PNN} \cite{PNN} & \textbf{2016} & Adapts a simple three-layer architecture for pansharpening.\\
\textbf{PanNet} \cite{PanNet} & \textbf{2017} & Deeper CNN for pansharpening, incorporating domain-specific knowledge to preserve both spectral and spatial information.\\
\textbf{DiCNN} \cite{DiCNN} & \textbf{2018} & Proposes a new detail injection-based convolutional neural network framework for pansharpening.\\
\textbf{FusionNet} \cite{FusionNet} & \textbf{2021} & Introduces the use of deep convolutional neural networks combined with traditional fusion schemes for pansharpening.\\
\textbf{DCFNet} \cite{DCFNet} & \textbf{2021} & Addresses the limitations of single-scale feature fusion by considering both high-level semantics and low-level features.\\
\textbf{LAGConv} \cite{Jin2022LAGConvLA} & \textbf{2022} & Employs local-context adaptive convolution kernels with global harmonic bias.\\
\textbf{HMPNet} \cite{HMPNet} & \textbf{2023} & An interpretable model-driven deep network for fusing hyperspectral, multispectral, and panchromatic images.\\
\textbf{CMT} \cite{CMT} & \textbf{2024} & Integrates a signal-processing-inspired modulation technique into the attention mechanism to effectively fuse images.\\
\textbf{CANNet} \cite{Duan2024ContentAdaptiveNC} & \textbf{2024} &  Incorporates non-local self-similarity to improve the effectiveness and reduce redundant learning in remote sensing image fusion. \\
\bottomrule 
\end{tabularx}

%% file: main.bbl
\begin{thebibliography}{10}

\bibitem{EXP}
Bruno Aiazzi, Luciano Alparone, Stefano Baronti, and Andrea Garzelli.
\newblock Context-driven fusion of high spatial and spectral resolution images based on oversampled multiresolution analysis.
\newblock {\em IEEE Trans. Geosci. Remote. Sens.}, 40:2300--2312, 2002.

\bibitem{Arienzo2022FullResolutionQA}
A.~Arienzo, Gemine Vivone, Andrea Garzelli, Luciano Alparone, and Jocelyn Chanussot.
\newblock Full-resolution quality assessment of pansharpening: Theoretical and hands-on approaches.
\newblock {\em IEEE Geoscience and Remote Sensing Magazine}, 10:168--201, 2022.

\bibitem{Boardman1993AutomatingSU}
Joseph~W. Boardman.
\newblock Automating spectral unmixing of aviris data using convex geometry concepts.
\newblock 1993.

\bibitem{CS1}
Jaewan Choi, Kiyun Yu, and Yongil Kim.
\newblock A new adaptive component-substitution-based satellite image fusion by using partial replacement.
\newblock {\em IEEE Transactions on Geoscience and Remote Sensing}, 49:295--309, 2011.

\bibitem{Dai2017DeformableCN}
Jifeng Dai, Haozhi Qi, Yuwen Xiong, Yi~Li, Guodong Zhang, Han Hu, and Yichen Wei.
\newblock Deformable convolutional networks.
\newblock {\em 2017 IEEE International Conference on Computer Vision (ICCV)}, pages 764--773, 2017.

\bibitem{FusionNet}
Liang-Jian Deng, Gemine Vivone, Cheng Jin, and Jocelyn Chanussot.
\newblock Detail injection-based deep convolutional neural networks for pansharpening.
\newblock {\em IEEE Transactions on Geoscience and Remote Sensing}, 59:6995--7010, 2021.

\bibitem{Deng2021DetailID}
Liang-Jian Deng, Gemine Vivone, Cheng Jin, and Jocelyn Chanussot.
\newblock Detail injection-based deep convolutional neural networks for pansharpening.
\newblock {\em IEEE Transactions on Geoscience and Remote Sensing}, 59:6995--7010, 2021.

\bibitem{Pancollection}
Liang-Jian Deng, Gemine Vivone, Mercedes~Eugenia Paoletti, Giuseppe Scarpa, Jiang He, Yongjun Zhang, Jocelyn Chanussot, and Antonio~J. Plaza.
\newblock Machine learning in pansharpening: A benchmark, from shallow to deep networks.
\newblock {\em IEEE Geoscience and Remote Sensing Magazine}, 10:279--315, 2022.

\bibitem{Duan2024ContentAdaptiveNC}
Yule Duan, Xiao Wu, Haoyu Deng, and Liangjian Deng.
\newblock Content-adaptive non-local convolution for remote sensing pansharpening.
\newblock In {\em Proceedings of the 2024 IEEE/CVF Conference on Computer Vision and Pattern Recognition (CVPR)}, pages 27738--27747, 2024.

\bibitem{PYconv}
Ionut~Cosmin Duta, Li~Liu, Fan Zhu, and Ling Shao.
\newblock Pyramidal convolution: Rethinking convolutional neural networks for visual recognition.
\newblock {\em ArXiv}, abs/2006.11538, 2020.

\bibitem{VO1}
Xueyang Fu, Zihuang Lin, Yue Huang, and Xinghao Ding.
\newblock A variational pan-sharpening with local gradient constraints.
\newblock {\em 2019 IEEE/CVF Conference on Computer Vision and Pattern Recognition (CVPR)}, pages 10257--10266, 2019.

\bibitem{CVPR19}
Xueyang Fu, Zihuang Lin, Yue Huang, and Xinghao Ding.
\newblock A variational pan-sharpening with local gradient constraints.
\newblock {\em 2019 IEEE/CVF Conference on Computer Vision and Pattern Recognition (CVPR)}, pages 10257--10266, 2019.

\bibitem{Garzelli2009HypercomplexQA}
Andrea Garzelli and Filippo Nencini.
\newblock Hypercomplex quality assessment of multi/hyperspectral images.
\newblock {\em IEEE Geoscience and Remote Sensing Letters}, 6:662--665, 2009.

\bibitem{He2015DeepRL}
Kaiming He, X.~Zhang, Shaoqing Ren, and Jian Sun.
\newblock Deep residual learning for image recognition.
\newblock {\em 2016 IEEE Conference on Computer Vision and Pattern Recognition (CVPR)}, pages 770--778, 2015.

\bibitem{DiCNN}
Lin He, Yizhou Rao, Jun~Yu Li, Jocelyn Chanussot, Antonio~J. Plaza, Jiawei Zhu, and Bo~Li.
\newblock Pansharpening via detail injection based convolutional neural networks.
\newblock {\em IEEE Journal of Selected Topics in Applied Earth Observations and Remote Sensing}, 12:1188--1204, 2018.

\bibitem{Jin2022LAGConvLA}
Zi-Rong Jin, Tian-Jing Zhang, Tai-Xiang Jiang, Gemine Vivone, and Liang-Jian Deng.
\newblock Lagconv: Local-context adaptive convolution kernels with global harmonic bias for pansharpening.
\newblock In {\em AAAI Conference on Artificial Intelligence}, 2022.

\bibitem{Kingma2014AdamAM}
Diederik~P. Kingma and Jimmy Ba.
\newblock Adam: A method for stochastic optimization.
\newblock {\em CoRR}, abs/1412.6980, 2014.

\bibitem{Li2019SelectiveKN}
Xiang Li, Wenhai Wang, Xiaolin Hu, and Jian Yang.
\newblock Selective kernel networks.
\newblock {\em 2019 IEEE/CVF Conference on Computer Vision and Pattern Recognition (CVPR)}, pages 510--519, 2019.

\bibitem{PNN}
Giuseppe Masi, Davide Cozzolino, Luisa Verdoliva, and Giuseppe Scarpa.
\newblock Pansharpening by convolutional neural networks.
\newblock {\em Remote. Sens.}, 8:594, 2016.

\bibitem{Meng2019ReviewOT}
Xiangchao Meng, Huanfeng Shen, Huifang Li, Liangpei Zhang, and Randi Fu.
\newblock Review of the pansharpening methods for remote sensing images based on the idea of meta-analysis: Practical discussion and challenges.
\newblock {\em Inf. Fusion}, 46:102--113, 2019.

\bibitem{TV}
Frosti Palsson, Johannes~R. Sveinsson, and Magnus~Orn Ulfarsson.
\newblock A new pansharpening algorithm based on total variation.
\newblock {\em IEEE Geoscience and Remote Sensing Letters}, 11:318--322, 2014.

\bibitem{snackConvolution}
Yaolei Qi, Yuting He, Xiaoming Qi, Yuanyuan Zhang, and Guanyu Yang.
\newblock Dynamic snake convolution based on topological geometric constraints for tubular structure segmentation.
\newblock {\em 2023 IEEE/CVF International Conference on Computer Vision (ICCV)}, pages 6047--6056, 2023.

\bibitem{Unet1}
Olaf Ronneberger, Philipp Fischer, and Thomas Brox.
\newblock U-net: Convolutional networks for biomedical image segmentation.
\newblock {\em ArXiv}, abs/1505.04597, 2015.

\bibitem{CMT}
Wenjie Shu, Hong-Xia Dou, Rui Wen, Xiao Wu, and Liang-Jian Deng.
\newblock Cmt: Cross modulation transformer with hybrid loss for pansharpening.
\newblock {\em IEEE Geoscience and Remote Sensing Letters}, 21:1--5, 2024.

\bibitem{Su2019PixelAdaptiveCN}
Hang Su, V.~Jampani, Deqing Sun, Orazio Gallo, Erik~G. Learned-Miller, and Jan Kautz.
\newblock Pixel-adaptive convolutional neural networks.
\newblock {\em 2019 IEEE/CVF Conference on Computer Vision and Pattern Recognition (CVPR)}, pages 11158--11167, 2019.

\bibitem{Tian2021VariationalPB}
Xin Tian, Yuerong Chen, Changcai Yang, and Jiayi Ma.
\newblock Variational pansharpening by exploiting cartoon-texture similarities.
\newblock {\em IEEE Transactions on Geoscience and Remote Sensing}, 60:1--16, 2021.

\bibitem{HMPNet}
Xin Tian, Kun Li, Wei Zhang, Zhongyuan Wang, and Jiayi Ma.
\newblock Interpretable model-driven deep network for hyperspectral, multispectral, and panchromatic image fusion.
\newblock {\em IEEE Transactions on Neural Networks and Learning Systems}, 35:14382--14395, 2023.

\bibitem{CS2}
Gemine Vivone.
\newblock Robust band-dependent spatial-detail approaches for panchromatic sharpening.
\newblock {\em IEEE Transactions on Geoscience and Remote Sensing}, 57:6421--6433, 2019.

\bibitem{BSDS-PC}
Gemine Vivone.
\newblock Robust band-dependent spatial-detail approaches for panchromatic sharpening.
\newblock {\em IEEE Transactions on Geoscience and Remote Sensing}, 57:6421--6433, 2019.

\bibitem{Vivone2018FullSR}
Gemine Vivone, Rocco Restaino, and Jocelyn Chanussot.
\newblock Full scale regression-based injection coefficients for panchromatic sharpening.
\newblock {\em IEEE Transactions on Image Processing}, 27:3418--3431, 2018.

\bibitem{MTF-GLP-FS}
Gemine Vivone, Rocco Restaino, and Jocelyn Chanussot.
\newblock Full scale regression-based injection coefficients for panchromatic sharpening.
\newblock {\em IEEE Transactions on Image Processing}, 27:3418--3431, 2018.

\bibitem{MRA1}
Gemine Vivone, Rocco Restaino, Mauro~Dalla Mura, Giorgio Licciardi, and Jocelyn Chanussot.
\newblock Contrast and error-based fusion schemes for multispectral image pansharpening.
\newblock {\em IEEE Geoscience and Remote Sensing Letters}, 11:930--934, 2014.

\bibitem{Wald2002DataFD}
Lucien Wald.
\newblock Data fusion. definitions and architectures - fusion of images of different spatial resolutions.
\newblock 2002.

\bibitem{Wald1997FusionOS}
Lucien Wald, Thierry Ranchin, and Marc Mangolini.
\newblock Fusion of satellite images of different spatial resolutions: Assessing the quality of resulting images.
\newblock {\em Photogrammetric Engineering and Remote Sensing}, 63:691--699, 1997.

\bibitem{Wang2021SSconvES}
Yudong Wang, Liang-Jian Deng, Tian-Jing Zhang, and Xiao Wu.
\newblock Ssconv: Explicit spectral-to-spatial convolution for pansharpening.
\newblock {\em Proceedings of the 29th ACM International Conference on Multimedia}, 2021.

\bibitem{DCFNet}
Xiao Wu, Tingzhu Huang, Liang-Jian Deng, and Tian-Jing Zhang.
\newblock Dynamic cross feature fusion for remote sensing pansharpening.
\newblock {\em 2021 IEEE/CVF International Conference on Computer Vision (ICCV)}, pages 14667--14676, 2021.

\bibitem{LRTCFPan}
Zhong-Cheng Wu, Ting-Zhu Huang, Liang-Jian Deng, Jie Huang, Jocelyn Chanussot, and Gemine Vivone.
\newblock Lrtcfpan: Low-rank tensor completion based framework for pansharpening.
\newblock {\em IEEE Transactions on Image Processing}, 32:1640--1655, 2023.

\bibitem{PanNet}
Junfeng Yang, Xueyang Fu, Yuwen Hu, Yue Huang, Xinghao Ding, and John Paisley.
\newblock Pannet: A deep network architecture for pan-sharpening.
\newblock In {\em 2017 IEEE International Conference on Computer Vision (ICCV)}, pages 1753--1761, 2017.

\bibitem{scale-adaptive}
Rui Zhang, Sheng Tang, Yongdong Zhang, Jintao Li, and Shuicheng Yan.
\newblock Scale-adaptive convolutions for scene parsing.
\newblock {\em 2017 IEEE International Conference on Computer Vision (ICCV)}, pages 2050--2058, 2017.

\bibitem{Zhou2021DecoupledDF}
Jingkai Zhou, V.~Jampani, Zhixiong Pi, Qiong Liu, and Ming-Hsuan Yang.
\newblock Decoupled dynamic filter networks.
\newblock {\em 2021 IEEE/CVF Conference on Computer Vision and Pattern Recognition (CVPR)}, pages 6643--6652, 2021.

\bibitem{Zhu2018DeformableCV}
Xizhou Zhu, Han Hu, Stephen Lin, and Jifeng Dai.
\newblock Deformable convnets v2: More deformable, better results.
\newblock {\em 2019 IEEE/CVF Conference on Computer Vision and Pattern Recognition (CVPR)}, pages 9300--9308, 2018.

\end{thebibliography}
